\documentclass[10pt,twocolumn,letterpaper]{article}
\usepackage[pagenumbers]{cvpr}
\usepackage{amsmath,amssymb,amsthm,booktabs,multirow,xspace,bm,adjustbox}
\definecolor{cvprblue}{rgb}{0.21,0.49,0.74}
\usepackage[pagebackref,breaklinks,colorlinks,allcolors=cvprblue]{hyperref}
\usepackage[capitalize]{cleveref}
\usepackage{placeins}
\crefname{section}{Sec.}{Secs.}\Crefname{section}{Section}{Sections}\crefname{table}{Tab.}{Tabs.}\Crefname{table}{Table}{Tables}
\newtheorem{definition}{Definition}\newtheorem{proposition}{Proposition}\newtheorem{lemma}{Lemma}
\newcommand{\ours}{ORDO\xspace}
\newcommand{\std}[1]{\textsubscript{\,$\pm$#1}}
\newcommand{\multiset}{\mathcal{M}}

\newcommand{\gainQnine}{+7.6}
\newcommand{\tomatoQnine}{+21.7}
\newcommand{\shufQnine}{+0.8}

\newcommand{\threeQnine}{+3.0}

\newcommand{\mvbQnine}{+2.1}

\newcommand{\gainQvl}{+4.9}
\newcommand{\tomatoQvl}{+13.8}
\newcommand{\shufQvl}{$-$0.8}

\newcommand{\threeQvl}{+2.0}

\newcommand{\mvbQvl}{+0.4}

\newcommand{\gainMolmo}{+5.1}
\newcommand{\tomatoMolmo}{+17.3}

\newcommand{\threeMolmo}{+1.1}
\newcommand{\threeShufMolmo}{$-$1.2}
\newcommand{\mvbMolmo}{$-$1.4}

\newcommand{\gainQfour}{+7.0}

\newcommand{\threeQfour}{+2.8}
\newcommand{\threeShufQfour}{+0.4}
\newcommand{\mvbQfour}{+2.6}

\newcommand{\mevalBase}{48.8}
\newcommand{\mevalOrdo}{81.3}
\newcommand{\mevalShuf}{52.0}
\newcommand{\thinkVsThink}{+6.5}
\newcommand{\placMean}{$-$5.3}

\newcommand{\veloBase}{15.7}
\newcommand{\oiMean}{+1.9}
\newcommand{\vloiMean}{$-$1.4}
\newcommand{\leakMeanThree}{+5.7}
\newcommand{\vlleakMeanTwo}{+4.4}

\title{ORDO: Fine-Grained Motion Understanding in Video LLMs\\via Order-Dependent Supervision}
\author{
Jiaye Luan\textsuperscript{1}\quad
Chaoning Zhang\textsuperscript{1,}\thanks{Corresponding author.}\quad
Pengcheng Zheng\textsuperscript{1}\quad
Dongshen Han\textsuperscript{2}\\
Sheng Zheng\textsuperscript{1}\quad
Ping Guo\textsuperscript{3}\quad
Caiyan Qin\textsuperscript{4}\quad
Yang Du\textsuperscript{5}\quad
Yang Yang\textsuperscript{1}\\[0.5em]
{\small\textsuperscript{1}University of Electronic Science and Technology of China}\\
{\small\textsuperscript{2}Kyung Hee University\quad
\textsuperscript{3}Hong Kong Yuwin Technology Co., Ltd.}\\
{\small\textsuperscript{4}Harbin Institute of Technology, Shenzhen\quad
\textsuperscript{5}Northeast Forest University}\\[0.4em]
{\small\texttt{jiaye.luan@std.uestc.edu.cn}\quad
\texttt{chaoningzhang1990@gmail.com}}
}
\begin{document}
\maketitle
\begin{abstract}
Video large language models (video LLMs) answer a wide range of questions about real videos, yet they remain weak at
fine-grained motion such as whether an object speeds up or slows down. Existing methods train these models on synthetic
motion clips, on order pretext tasks such as reordering shuffled clips, or with rewards that compare ordered and
perturbed frames. None of them requires that the label be unrecoverable from the frames once their order is discarded,
so a model can fit such data from static cues without learning to use frame order. We propose ORDO, a
framework that builds this requirement into the training data. We define order-necessary supervision as data in which
the unordered set of frames has the same distribution under competing labels, and we realise it for motion attributes
with three constructions, time-reversal pairing, matched spatial extent and uniform interval coverage. A procedural
curriculum of six motion primitives instantiates these constructions, and low-rank adapters of the language model are
trained on it with an answer-level objective and self-distillation replay on real videos. With 4{,}000 synthetic clips,
ORDO raises the mean accuracy over four motion benchmarks by \gainQnine{}, \gainQvl{} and \gainMolmo{} points on
Qwen3.5-9B, Qwen3-VL-8B and Molmo2-8B, while the same training with shuffled frames changes it by less than one point.
Code and generator will be made publicly available.
\end{abstract}

\section{Introduction}
\label{sec:intro}

\begin{figure}[t]
  \centering
  \includegraphics[width=\linewidth]{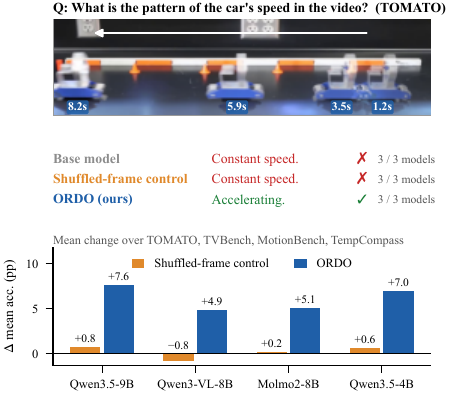}
  \caption{\textbf{Fine-grained motion requires frame order.} Top: a TOMATO question, in which the toy car at four
  equally spaced sampled times covers more distance in each interval. The three base models and their shuffled-frame
  controls answer ``constant speed'', and \ours answers correctly on all three backbones. Bottom: change in the mean
  accuracy over four motion benchmarks.}
  \label{fig:teaser}
\end{figure}

Video large language models (video LLMs) now answer a wide range of questions about real videos, but questions about
how things move remain difficult. On the velocity and frequency questions of TOMATO~\cite{shangguan2025tomato}, the
three recent models we study reach 15.7\%, 32.4\% and 30.5\%, against 24.4\% expected from guessing, and rotation
questions fare little better at 23.4 to 25.2\% against 20.0\%. A single frame cannot convey fine-grained motion, so
these numbers indicate that current models make limited use of frame order. In the example of \cref{fig:teaser}, all three base
models describe an accelerating car as moving at constant speed.

Several lines of work aim to strengthen temporal understanding through training. Synthetic data offers full control
over motion labels. SynRL renders temporal primitives with code and trains with reinforcement learning~\cite{jiang2026learning},
and MoDirect-Inst builds instruction data for signed motion
direction~\cite{lee2026which}. Other methods use frame order itself as the training signal. Visual Jigsaw and
VideoSSR train models to reorder shuffled clips~\cite{wu2025visualjigsaw,he2025videossr}, and ArrowRL rewards answers
on forward video that differ from those on the reversed video~\cite{xue2025seeing}. T-GRPO and TGPO reward answering better
from ordered than from shuffled frames~\cite{feng2025videor1,xu2026tgpo}.
Inference-time methods such as TAI instead re-inject temporal features into the language model without
training~\cite{shin2026before}.

These methods share an implicit assumption that motion labels or order-based objectives make a model use frame order.
The assumption is not checked on the training data. A model learns to use temporal order only to the extent that its
training signal depends on order, and much of popular video-language data can be fitted from static
content~\cite{buch2022revisiting,lei2023revealing}. Motion labels are no exception. When a clip that speeds up is
shorter, ends at a different position or covers a different region than one that slows down, the label can be read
from the frames as a set, and a model can fit it without ordering them. A recent controlled study finds that much of
the gain from reinforcement learning on synthetic temporal data indeed does not depend on the
frames~\cite{sadhu2026what}. Order pretext tasks have the opposite problem, since
they require order but say little about motion attributes. Benchmarks are built so that a single frame
cannot answer them~\cite{shangguan2025tomato,cores2025lost,liu2024tempcompass}, but training data is rarely held to
the same standard. What is missing is a criterion that states when a motion label
requires frame order, together with a way to construct training data that meets it at scale.

To address this, we propose \ours, a framework for order-dependent supervision. We call supervision
\emph{order-necessary} when the unordered set of frames has the same distribution under two labels, so that a model that
sees the frames without their order cannot separate them. Three constructions produce such supervision for motion
attributes. Time-reversal pairing gives each clip a partner label under playback reversal, matched spatial extent gives
labels without a partner the same trajectory endpoints and path, and uniform interval coverage removes the count from
the positions of a repeated motion. A procedural curriculum of six motion primitives instantiates the constructions, and
low-rank adapters of the language model are trained on it with an answer-level objective and self-distillation replay
on real videos. With 4{,}000 synthetic clips, \ours raises the mean accuracy over TOMATO, TVBench~\cite{cores2025lost},
MotionBench~\cite{hong2025motionbench} and TempCompass~\cite{liu2024tempcompass} by \gainQnine{} points on Qwen3.5-9B,
\gainQvl{} on Qwen3-VL-8B and \gainMolmo{} on Molmo2-8B, with the largest gains of 14 to 22 points on TOMATO. The
same training with shuffled frames changes the mean by $-0.8$ to $+0.8$ points, and the trained models lose their
advantage when their test frames are shuffled or reversed. \ours also outperforms prior synthetic motion data, order
pretext tasks, an inference-time method and test-time reasoning.

Our contributions are as follows.
\begin{itemize}
  \item We propose \ours, a framework that makes frame order necessary for the training labels of a video LLM. It
  formalises order-necessary supervision as a criterion on the data distribution, proves that a frame-shuffled control
  receives no signal for such labels, and trains low-rank adapters with an answer-level objective and
  self-distillation replay.
  \item We design three constructions that realise the criterion for motion attributes, time-reversal pairing, which
  satisfies it exactly, matched spatial extent and uniform interval coverage, and a procedural curriculum of six motion primitives built from them.
  The constructions act on time rather than content and also turn unlabelled real footage into order-necessary
  supervision.
  \item Experiments on three video LLMs and two scales raise the mean over four motion benchmarks by 4.9 to 7.6 points, gains that
  training-time and test-time controls attribute to frame order, and the gain grows with the share of order-necessary labels. Layer-wise
  probing identifies two regimes, in which training either makes an existing motion signal readable at the answer
  position or forms a signal that the base model does not encode linearly.
\end{itemize}

\section{Related work}
\label{sec:related}

\paragraph{Training video LLMs for temporal understanding.}
Rendered video supports controlled studies of temporal reasoning~\cite{yi2020clevrer,greff2022kubric}, and several
recent methods train video LLMs on synthetic motion data. SynRL renders
temporal primitives with code and trains Qwen2.5-VL and Qwen3-VL with chain-of-thought fine-tuning and reinforcement
learning~\cite{jiang2026learning}. MoDirect-Inst and DeltaDirect address signed motion direction with synthetic
instruction data and a projector-level motion loss~\cite{lee2026which}, SYNCR adapts Qwen3-VL with low-rank updates on
simulator-grounded cross-video tasks and uses a wrong-label adapter as a control~\cite{ghazanfari2026syncr}, and
MotionSight adds visual prompts and motion question-answer data~\cite{du2026motionsight}. A second group uses frame
order directly. The arrow of time is visually detectable~\cite{pickup2014seeing} and serves, like frame order, as a
self-supervised pretext task~\cite{misra2016shuffle,fernando2017self,wei2018learning}. For video LLMs, ArrowRL rewards
answers on forward video that differ from those on reversed video~\cite{xue2025seeing}, T-GRPO and TGPO reward
answering better from ordered than from shuffled frames~\cite{feng2025videor1,xu2026tgpo}, and CRPO rewards answers that
change under reversal or flipping only when the queried property changes~\cite{du2026crpo}. Visual Jigsaw and VideoSSR
train models to reorder shuffled clips among other tasks~\cite{wu2025visualjigsaw,he2025videossr}, while TimeWarp,
TEMPLE and VT-Contrast contrast ordered with perturbed clips~\cite{vani2025timewarp,li2026temple,shi2026shape}, as TACT
and RTime do for video-text retrieval~\cite{bagad2023test,du2024reversed}. These methods use order as a pretext, a
reward or a negative. Retro-Actions introduced label swapping for actions whose time reversal is another action
class~\cite{price2019retro}, and our time-reversal construction applies this idea to motion attributes. A controlled
study of verifiable rewards on synthetic clips finds that much of the gain does not depend on the frames
and recommends frame-free controls~\cite{sadhu2026what}. We turn this observation into a criterion that training data
can be built to satisfy.

\paragraph{Temporal information inside video LLMs.}
Several analyses locate temporal information in video LLMs. They report cross-frame interaction in early and middle
layers~\cite{kim2026map}, a bottleneck at the vision-language projector~\cite{han2026tracing}, motion direction that is
linearly decodable but not reflected in answers~\cite{lee2026which}, repetition counts that are decodable when the
model reports them incorrectly~\cite{khoshnoodi2026repprobe}, and ego-motion cues that probes and light fine-tuning
recover in driving videos~\cite{sreeram2026mind}. Forward and reverse representations also differ less toward the
output layers, which TAI counters at inference~\cite{shin2026before}, and rebalancing decoder attention away from a
dominant anchor frame reduces hallucination~\cite{liu2026relaxing}. Our analysis extends these
observations to speed change on three recent backbones, identifies a second regime in which training makes decodable a
signal that the base model does not encode linearly, and tests with subspace ablation whether the model relies on that direction.

\section{Method}
\label{sec:method}

\begin{figure*}[t]
  \centering
  \includegraphics[width=\linewidth]{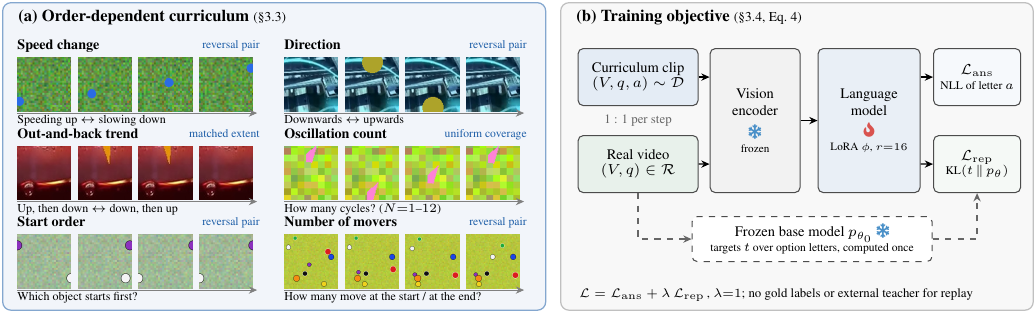}
  \caption{\textbf{Overview of \ours.} (a) The curriculum consists of six motion primitives, shown with generator frames
  and the construction that makes each label order-dependent. (b) The answer loss on curriculum clips trains low-rank
  adapters of the language model, and a replay loss on real videos keeps the option distribution close to that of the
  frozen base model.}
  \label{fig:overview}
\end{figure*}

\subsection{Overview}
\label{sec:overview}

A video LLM $p_\theta(y \mid V, q)$ answers a multiple-choice question $q$ about a clip $V = (x_1, \dots, x_T)$.
\cref{fig:overview} summarises \ours. \cref{sec:def} states when a label requires frame order and shows that a
model without access to the order cannot learn such a label. \cref{sec:curriculum} gives three constructions that
produce such labels for motion attributes and assembles them into a procedural curriculum, and \cref{sec:objective}
trains low-rank adapters on it and defines the controls that measure how much of the gain depends on order.

\subsection{Order-necessary supervision}
\label{sec:def}

For a permutation $\pi$ of $\{1, \dots, T\}$ let $\pi V = (x_{\pi(1)}, \dots, x_{\pi(T)})$, and let
$\multiset(V) = \{\!\{x_1, \dots, x_T\}\!\}$ be the multiset of frames, which is the information a clip retains once its
order is discarded. A labelling function $f$ is \emph{order-invariant} if $f(\pi V) = f(V)$ for all $V$ and $\pi$, or
equivalently if $f$ factors through $\multiset$. Questions about which objects, scenes or attributes appear are
order-invariant, and many video-language benchmarks and training sets can be solved to a large extent from a single
frame~\cite{buch2022revisiting,lei2023revealing}. A model can fit such data without representing order, so we
characterise supervision at the level of the data distribution. The number of frames $T$ is a property of the clip and
is visible in its multiset. Frame times are attached to positions in the sequence rather than to frames, so they carry
no information beyond $T$.

\begin{definition}[Order-necessary supervision]
\label{def:on}
Let $\mathcal{D}$ be a distribution over clips and labels, and let $c \neq c'$ be two labels. $\mathcal{D}$ is
\emph{order-necessary for $(c, c')$} if, for every question $q$, the frame multiset has the same distribution under
both labels,
\begin{equation}
  P_{\mathcal{D}}\bigl(\multiset(V) \mid q, y = c\bigr) = P_{\mathcal{D}}\bigl(\multiset(V) \mid q, y = c'\bigr).
  \label{eq:on}
\end{equation}
\end{definition}

\begin{proposition}
\label{prop:bound}
If $\mathcal{D}$ is order-necessary for $(c, c')$, every predictor $g$ of the multiset satisfies
$P_{\mathcal{D}}\bigl(g(\multiset(V), q) = y \mid q, y \in \{c, c'\}\bigr) \le \max\bigl(P(y{=}c \mid q), P(y{=}c' \mid q)\bigr) / P(y \in \{c, c'\} \mid q)$.
The same bound holds for any predictor that receives the frames in a uniformly random order.
\end{proposition}

\begin{proof}
Conditioned on $q$ and $y \in \{c, c'\}$, \cref{eq:on} makes $\multiset(V)$ independent of $y$, so the posterior of $y$ given
$\multiset(V)$ equals its prior and no function of $\multiset(V)$ exceeds the majority rate. Given $\multiset(V)$, a
uniformly permuted clip $\pi V$ is distributed independently of $y$. A predictor of $\pi V$ is therefore a randomised
predictor of $\multiset(V)$, to which the same argument applies.
\end{proof}

The condition is sufficient, and it makes the criterion testable. Training on frame-shuffled clips provides no
population-level signal for separating an order-necessary label pair, so a model trained on shuffled frames measures how
much of a gain depends on order. Order necessity is stronger than per-example order sensitivity,
$\exists \pi : f(\pi V) \neq f(V)$, which holds for most motion questions even when static cues in the training
distribution decide the answer, so a criterion on the distribution is needed.

\subsection{Constructing an order-dependent curriculum}
\label{sec:curriculum}

\begin{figure}[t]
  \centering
  \includegraphics[width=\linewidth]{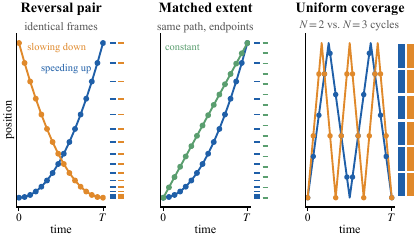}
  \caption{\textbf{The three constructions} as position over time. Dots are sampled frames, and the ticks at the right of
  each panel show the unordered set of positions. A reversal pair has identical sets under opposite labels, matched
  extent equalises endpoints and path length, and uniform coverage makes the position distribution nearly independent of
  the count.}
  \label{fig:constructions}
\end{figure}

Each of our three constructions specifies how the generator samples trajectories, and \cref{fig:constructions} shows
the positions they produce over time.

\paragraph{Three constructions.}
Let $\rho V = (x_T, \dots, x_1)$ denote time reversal. Many motion attributes map to a partner attribute under $\rho$,
such as speeding up and slowing down, left and right, approaching and receding, and which of two objects starts moving
first when both move for the same duration. Time-reversal pairing exploits this symmetry.

\begin{lemma}
\label{lem:rev}
Let $\sigma$ be an involution on labels and $c$ a label with $\sigma(c) \neq c$. If, for every question $q$, the
generator satisfies $V \mid (q, y{=}c) \overset{d}{=} \rho V \mid (q, y{=}\sigma(c))$, then $\mathcal{D}$ is
order-necessary for $(c, \sigma(c))$.
\end{lemma}

\begin{proof}
$\multiset(\rho V) = \multiset(V)$ for every clip, so the two conditional distributions of the multiset coincide.
\end{proof}

The generator meets this condition. Endpoints are drawn symmetrically, each partner profile is the time reversal of
the other, and appearance, distractors and frame times are independent of the label. For the number of movers,
reversal maps a start question to an end question, and the start and end counts are drawn exchangeably, so the
frame multiset does not reveal which count is which. Labels without a reversal partner, namely constant
speed, one-way motion and simultaneous onset, receive trajectories by \emph{matched spatial extent}. Their endpoints,
path length and covered interval match those of the paired labels, so only the spacing of sampled positions differs,
and the shuffled-frame control measures how much separability remains. For counting repeated motion we use
\emph{uniform interval coverage}. The clip duration is independent of the count, and the object follows a
constant-speed triangle wave over a fixed interval, so the sampled positions are near-uniform on that interval for
every count and the frame set carries almost no information about the number of cycles.

\paragraph{Curriculum.}
We instantiate the constructions as six motion primitives, shown in \cref{fig:overview}a with their training
proportions given below. In \textbf{speed change} (34\%) an object translates, rotates or oscillates, or the camera pans
across a still. Speeding up and slowing down form a reversal pair, and constant speed uses matched extent.
\textbf{Direction} (13\%) covers translation, camera panning, and approach and recession rendered by scale change, with
reversal pairs left and right, up and down, and toward and away. \textbf{Out-and-back trend} (13\%) moves an object along
one axis, either returning (``left, then right'') or travelling one way. \textbf{Oscillation count} (16\%) shows
$N \in \{1, \dots, 12\}$ back-and-forth cycles in clips of 30 to 32\,s. The options are $N$ and three to five
neighbouring values, with the rank of $N$ drawn uniformly so that option values carry no cue. In \textbf{start order}
(12\%) two objects each move once for the same duration, and the question asks which starts first. In \textbf{number of
movers} (12\%) each clip has two to eight objects, some of which move throughout the clip, near its start or near its
end, and $\rho$ exchanges start and end questions. In 30\% of the speed, direction, trend and count clips the target is a
named coloured ball among one to three other moving balls. Backgrounds are stills from Video-MME~\cite{fu2025videomme}
videos or procedural textures, and clips are rendered at 2 frames per second. The held-out test split uses different
backgrounds and question templates. Training uses 4{,}000 clips, and the appendix describes the generator
in full. We chose the primitives with the four motion benchmarks in view, adding counting for object and repetition
counting, whereas MVBench, Vinoground, TimeBlind and the test-time order controls played no part in any design
decision.

The constructions act on time rather than on what is rendered, so they also apply to unlabelled real video. In 12\,s
windows of steadily moving footage from Video-MME, split by video, we re-time playback with the speed profiles of the
curriculum and play the footage forwards or backwards independently of the label. A reversed speeding-up clip is then
distributed exactly like a slowing-down one, as \cref{lem:rev} requires. Playing a window back and forth $N$ times over a fixed
24\,s gives uniform coverage, and forward and reversed playback form a reversal pair.

\subsection{Training objective and order controls}
\label{sec:objective}

Let $\theta_0$ be the pretrained model. We adapt the language model with low-rank updates~\cite{hu2022lora}
$\theta = \theta_0 + \Delta(\phi)$ on all attention, linear-attention and MLP projections with rank $16$, and keep the
vision encoder and the projector frozen. For a curriculum example $(V, q, a)$ with answer letter $a$, the answer loss is
the negative log-likelihood of the letter and the end-of-turn token,
\begin{equation}
  \mathcal{L}_{\mathrm{ans}}(\phi) = -\log p_\theta(a \mid V, q) - \log p_\theta(\langle\mathrm{eot}\rangle \mid V, q, a).
\end{equation}
Training only on synthetic clips shifts the answer distribution of the model on real videos, which we counter with
self-distillation replay. The replay set $\mathcal{R}$ contains 2{,}140 real videos with multiple-choice questions from
Video-MME~\cite{fu2025videomme} and LVBench~\cite{wang2025lvbench}, neither of which is among our evaluation
benchmarks, and their gold answers are not used. For each item the base model provides a target distribution
$t = p_{\theta_0}(\cdot \mid V, q)$ over the option letters, computed once before training, and the replay loss is
\begin{equation}
  \mathcal{L}_{\mathrm{rep}}(\phi) = \mathrm{KL}\bigl(t \,\big\|\, p_\theta(\cdot \mid V, q)\bigr),
\end{equation}
with both distributions restricted to the option letters. Each optimisation step pairs every curriculum example with a
replay example and minimises
\begin{equation}
  \mathcal{L}(\phi) = \mathbb{E}_{(V,q,a) \sim \mathcal{D}}\,\mathcal{L}_{\mathrm{ans}}
  + \lambda\, \mathbb{E}_{(V,q) \sim \mathcal{R}}\,\mathcal{L}_{\mathrm{rep}}, \qquad \lambda = 1.
  \label{eq:objective}
\end{equation}
The replay term anchors each backbone to its own predictions on real video, as in distillation-based
continual learning~\cite{hinton2015distilling,li2018learning}, without labels or an external teacher.

\label{sec:controls}
Two controls separate the contribution of temporal order from other effects of training. The \emph{shuffled-frame
control} trains on the same clips, labels, replay data and schedule as \ours, with an independent random frame
permutation for each curriculum clip. By \cref{prop:bound} this model cannot learn to distinguish reversal-paired
labels, although it receives the same appearance statistics, answer formats, label priors and replay signal. Because
frame times follow sequence positions, a permuted clip carries no timing information about its original order. The
\emph{random-label control} keeps the frames and their order, replaces every answer with a uniformly drawn option, and
isolates the effect of the training procedure and of exposure to the clips. Both controls are evaluated on ordered
frames, and we take the gain of \ours over its shuffled-frame control as the part of the improvement that depends on
temporal order. On the held-out curriculum test split, the shuffled-frame control of Qwen3.5-9B reaches \mevalShuf\%
against \mevalBase\% for the base model and \mevalOrdo\% for \ours (seed 0), which is consistent with the construction
and shows that the residual separability of the matched-extent labels is small.

\section{Experiments}
\label{sec:exp}

\subsection{Implementation details}
\label{sec:setup}

\paragraph{Benchmarks and metrics.}
Four benchmarks target motion and temporal order, namely TOMATO~\cite{shangguan2025tomato} with 1{,}484 questions,
TVBench~\cite{cores2025lost} with 2{,}205, MotionBench~\cite{hong2025motionbench} with 2{,}000 development questions and
TempCompass~\cite{liu2024tempcompass} with 1{,}580 multiple-choice questions, and the appendix describes the subsets.
MVBench~\cite{li2024mvbench}, a broad 20-task benchmark, measures video understanding beyond motion. Two benchmarks of video
minimal pairs, TimeBlind~\cite{li2026timeblind} and Vinoground~\cite{zhang2024vinoground}, were used in no design
decision and serve as held-out evaluations. None of these benchmarks supplies training or replay data. We report
accuracy changes from the base model on identical items, with 95\% paired-bootstrap intervals~\cite{efron1993introduction}
and standard deviations across training seeds.

\paragraph{Backbones and training configurations.}
We train \ours on three recent open video LLMs with different language-model architectures and visual tokenisation,
Qwen3.5-9B~\cite{qwenteam2026qwen35} in its non-thinking mode, Qwen3-VL-8B-Instruct~\cite{bai2025qwen3vl} and
Molmo2-8B~\cite{clark2026molmo2}, and add Qwen3.5-4B as a second scale. All share the curriculum, objective and
hyper-parameters, and each uses replay targets from its own base model. Training uses 500 AdamW steps of 8 clips with
learning rate $10^{-4}$, cosine decay and LoRA rank 16 ($\alpha = 32$). All models are evaluated with
vLLM~\cite{kwon2023efficient} under one protocol, with frames sampled uniformly at 2 frames per second up to 64 frames,
the longer side resized to at most 448 pixels, greedy decoding and an instruction to answer with the option letter.

\paragraph{Comparison methods.}
Besides the order controls of \cref{sec:objective}, we compare against two kinds of alternative. The first replaces our
curriculum with 4{,}000 examples of other training data under our protocol and replay, namely
SynRL clips from the official generator~\cite{jiang2026learning}, rendered with letter answers at 4 frames per second to
reduce aliasing, MoDirect-Inst~\cite{lee2026which}, and two order pretext tasks on our own clips, a temporal jigsaw that
asks which of six orderings of three segments is shown and forward-versus-reversed
classification~\cite{pickup2014seeing,wei2018learning} of the real-footage windows, a task ArrowRL also trains
on~\cite{xue2025seeing}. The second group needs no training data. TAI~\cite{shin2026before} re-injects temporal
features at inference and runs in its reference implementation against its own baseline, and thinking mode lets
Qwen3.5-9B reason with a budget of 16k tokens before answering.

\subsection{Main results}
\label{sec:main}

\begin{table}[t]
  \centering
  \small
  \caption{\textbf{Main results.} Base rows give accuracy (\%), and other rows give the change in accuracy (points)
  from the base model on identical items. \ours rows show the mean and standard deviation across training seeds. Mean
  is over the four benchmarks, and w/o TOM.\ is the mean over TVBench, MotionBench and TempCompass. The controls use the
  same clips, replay data and schedule, with the frames of each training clip permuted (shuffled-frame) or each answer
  drawn uniformly (random-label), and they are evaluated on ordered frames. Per-benchmark 95\% confidence intervals are
  in the appendix.}
  \label{tab:main}
  \setlength{\tabcolsep}{4pt}
  \adjustbox{max width=\linewidth}{\begin{tabular}{@{}lcccccc@{}}
    \toprule
    & TOMATO & TVBench & MotionB. & TempC. & Mean & w/o TOM. \\
    \midrule
    \multicolumn{7}{l}{\textit{Qwen3.5-9B}} \\
\quad Base model & 32.5 & 56.1 & 62.9 & 75.1 & 56.6 & 64.7 \\
\quad Shuffled-frame control & +4.3 & $-$1.1 & $-$1.0 & +0.8 & +0.8 & $-$0.4 \\
\quad Random-label control & $-$7.5 & $-$7.3 & $-$1.8 & $-$4.6 & $-$5.3 & $-$4.6 \\
\quad \ours{} (3 seeds) & \textbf{+21.7\std{0.7}} & \textbf{+5.6\std{0.8}} & \textbf{$-$2.3\std{0.7}} & \textbf{+5.6\std{0.5}} & \textbf{+7.6\std{0.3}} & \textbf{+3.0\std{0.6}} \\
\midrule
\multicolumn{7}{l}{\textit{Qwen3-VL-8B}} \\
\quad Base model & 34.4 & 53.9 & 57.0 & 73.2 & 54.6 & 61.4 \\
\quad Shuffled-frame control & $-$0.6 & $-$2.4 & $-$0.2 & +0.1 & $-$0.8 & $-$0.8 \\
\quad \ours{} (3 seeds) & \textbf{+13.8\std{0.0}} & \textbf{+4.2\std{0.6}} & \textbf{$-$1.3\std{0.2}} & \textbf{+3.0\std{0.1}} & \textbf{+4.9\std{0.1}} & \textbf{+2.0\std{0.2}} \\
\midrule
\multicolumn{7}{l}{\textit{Molmo2-8B}} \\
\quad Base model & 36.4 & 56.9 & 60.7 & 72.1 & 56.5 & 63.2 \\
\quad Shuffled-frame control & +4.4 & $-$1.0 & $-$2.3 & $-$0.4 & +0.2 & $-$1.2 \\
\quad \ours{} (2 seeds) & \textbf{+17.3\std{0.9}} & \textbf{+3.1\std{1.1}} & \textbf{$-$0.5\std{0.1}} & \textbf{+0.6\std{0.9}} & \textbf{+5.1\std{0.2}} & \textbf{+1.1\std{0.0}} \\
\midrule
\multicolumn{7}{l}{\textit{Qwen3.5-4B}} \\
\quad Base model & 33.9 & 51.6 & 59.3 & 73.5 & 54.6 & 61.5 \\
\quad Shuffled-frame control & +1.1 & $-$0.4 & +0.4 & +1.3 & +0.6 & +0.4 \\
\quad \ours{} (1 seed) & \textbf{+19.4} & \textbf{+6.9} & \textbf{$-$3.1} & \textbf{+4.7} & \textbf{+7.0} & \textbf{+2.8} \\
 
    \bottomrule
  \end{tabular}}
\end{table}

\paragraph{Motion benchmarks.}
As \cref{tab:main} shows, \ours improves the four-benchmark mean of every backbone, by \gainQnine{} points for
Qwen3.5-9B, \gainQvl{} for Qwen3-VL-8B, \gainMolmo{} for Molmo2-8B and \gainQfour{} for Qwen3.5-4B. The largest change
is on TOMATO, whose questions were designed to resist single-frame solutions, with gains of \tomatoQnine{},
\tomatoQvl{} and \tomatoMolmo{} points. TVBench improves on every backbone and TempCompass on all but one Molmo2 seed,
and the three-benchmark means without TOMATO rise by \threeQnine{}, \threeQvl{}, \threeMolmo{} and \threeQfour{}
points. MotionBench changes by $-0.5$ to $-3.1$ points, and the seed standard deviation of the mean is at most 0.3. The shuffled-frame controls change the mean by only \shufQvl{} to \shufQnine{} points, and by
\threeShufMolmo{} to \threeShufQfour{} points without TOMATO. On TOMATO velocity both the control and \ours answer
``accelerating'' or ``decelerating'' far more often than the base model, in 66\% and 67\% of the 210 questions against
9\%, but only \ours tells the two apart. When it chooses one of them it is right in 96\% of cases with similar class
recalls of 75\% and 70\%, whereas the control is right in 55\% of cases against 50\% for
balanced guessing. The random-label control changes the mean by \placMean{} points, so exposure to the clips and the
training procedure do not produce the gain.

\begin{figure*}[t]
  \centering
  \includegraphics[width=\linewidth]{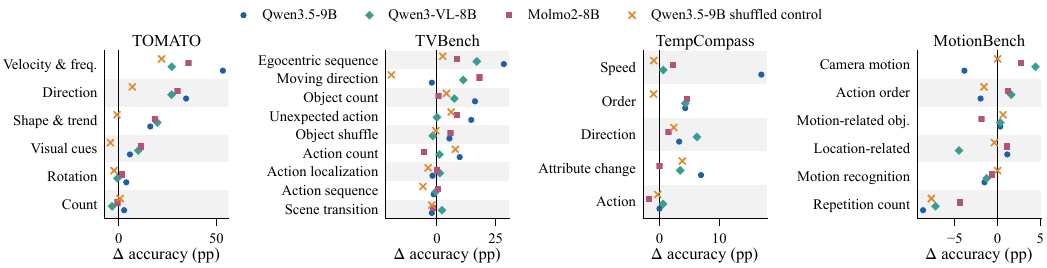}
  \caption{\textbf{Per-category change in accuracy} (points, with respect to each base model) for \ours (seed 0) on the
  three 8--9B backbones and for the shuffled-frame control of Qwen3.5-9B. Categories are sorted by the mean gain of the
  three backbones.}
  \label{fig:categories}
\end{figure*}

\paragraph{Where the gains arise.}
The per-category changes in \cref{fig:categories} concentrate where the answer depends on motion over time. They
include TOMATO velocity and frequency, where the base accuracy is \veloBase\% and Qwen3.5-9B gains 53.3 points, TOMATO
direction with gains of 27.0 to 34.5 points, TOMATO shape and trend, TVBench egocentric sequence, and TempCompass order
and direction, all on videos that the curriculum does not render. Appearance categories such as TempCompass action and
TVBench scene transition change by less than 2.5 points. MotionBench repetition counting declines by 8.7, 7.3 and 4.4
points, as it does by 7.8 points for the shuffled control, so the decline is not specific to order.

\begin{table}[t]
  \centering
  \small
  \caption{\textbf{Comparison with alternative supervision and inference-time methods} (change in accuracy, points,
  with respect to the base model, one seed). Data rows use our training protocol, including replay, with 4{,}000
  examples. $^\dagger$Evaluated in the reference implementation against its own baseline (applied by us to Qwen3.5 and
  Molmo2), with TOMATO on the 1{,}385 questions with unique keys. $^\ddagger$Thinking budget of 16k tokens, about 1.9k
  to 4.8k tokens per answer on average against 3 to 4 for the other rows. Chains over the budget keep the
  non-thinking answer. $^\S$Two-seed mean.}
  \label{tab:compare}
  \setlength{\tabcolsep}{3.2pt}
  \adjustbox{max width=\linewidth}{\begin{tabular}{@{}lccccc@{}}
    \toprule
    & TOMATO & TVB. & MoB. & TempC. & Mean \\
    \midrule
    \multicolumn{6}{l}{\textit{Qwen3.5-9B}} \\
\quad TAI~\cite{shin2026before}$^\dagger$ & +1.4 & +1.4 & $-$4.3 & +1.8 & +0.1 \\
\quad Thinking (16k)$^\ddagger$ & +6.8 & +5.7 & $-$2.6 & +3.4 & +3.3 \\
\quad SynRL data~\cite{jiang2026learning} & +7.3 & +2.0 & $-$0.4 & +0.5 & +2.4 \\
\quad MoDirect-Inst~\cite{lee2026which} & +5.3 & +2.8 & +0.5 & +1.1 & +2.4 \\
\quad Temporal jigsaw$^\S$ & +8.0 & +2.1 & +0.2 & +1.8 & +3.0 \\
\quad Arrow of time & +0.3 & +0.7 & $-$0.6 & +0.1 & +0.1 \\
\quad \ours{} & \textbf{+20.9} & \textbf{+6.3} & \textbf{$-$1.9} & \textbf{+6.2} & \textbf{+7.9} \\
\midrule\multicolumn{6}{l}{\textit{Qwen3-VL-8B}} \\
\quad TAI~\cite{shin2026before}$^\dagger$ & +2.8 & $-$0.1 & $-$1.9 & $-$0.5 & +0.1 \\
\quad SynRL data~\cite{jiang2026learning} & +0.9 & +0.9 & +0.7 & +1.1 & +0.9 \\
\quad Temporal jigsaw$^\S$ & +7.5 & +1.5 & +2.1 & +0.4 & +2.9 \\
\quad Arrow of time & +0.3 & $-$0.4 & +0.3 & +0.4 & +0.2 \\
\quad \ours{} & \textbf{+13.8} & \textbf{+3.5} & \textbf{$-$1.2} & \textbf{+3.0} & \textbf{+4.8} \\
\midrule\multicolumn{6}{l}{\textit{Molmo2-8B}} \\
\quad TAI~\cite{shin2026before}$^\dagger$ & +3.1 & +3.6 & $-$0.7 & +1.3 & +1.8 \\
\quad \ours{} & \textbf{+16.7} & \textbf{+2.3} & \textbf{$-$0.5} & \textbf{+1.3} & \textbf{+4.9} \\
 
    \bottomrule
  \end{tabular}}
\end{table}

\paragraph{Comparison with other sources of temporal improvement.}
\label{sec:compare}
The data rows of \cref{tab:compare} compare training data rather than systems. SynRL and MoDirect-Inst reach +2.4 points each on Qwen3.5-9B, against +7.9 for \ours at the same seed, and
SynRL reaches +0.9 against +4.8 on Qwen3-VL-8B. The gap persists on the targeted categories, where MoDirect-Inst
improves TOMATO direction by 12.2 points and \ours by 34.5. The shuffled-frame control attributes 2.4 points of the
SynRL gain to order, against 7.1 for \ours, and an order-free summary of the frames predicts the SynRL acceleration
labels, which is the situation that \cref{def:on} excludes. The order pretexts do not close the gap either. The temporal
jigsaw gains 3.0 and 2.9 points and the arrow-of-time task 0.1 and 0.2, with a training loss that stays at chance, so
an order pretext does not replace order-necessary labels. TAI changes the mean by +0.1, +0.1 and +1.8 points, and by less than one point at its original 16 frames. Thinking
adds 3.3 points to Qwen3.5-9B, and with thinking enabled for both models \ours still leads by \thinkVsThink{}.

\paragraph{Held-out minimal pairs and broad video understanding.}
On TimeBlind, where an instance counts only if all four video and question combinations are answered correctly with a
chance level of 6.25\%, \ours raises instance accuracy from 25.0\% to 42.7\% for Qwen3.5-9B as a mean of three seeds and
from 20.2\% to 33.3\% for Qwen3-VL-8B. On Vinoground it raises the text score of Qwen3.5-9B from 76.2\% to 83.9\% and that
of Qwen3-VL-8B from 67.6\% to 73.2\%. With shuffled test frames all models fall to 4 to 11\% on TimeBlind and near
chance on Vinoground. On MVBench, \ours changes accuracy by \mvbQnine{}, \mvbQvl{}, \mvbMolmo{} and \mvbQfour{} points
for Qwen3.5-9B, Qwen3-VL-8B, Molmo2-8B and Qwen3.5-4B.

\subsection{Ablation studies}
\label{sec:ablate}

\begin{table}[t]
  \centering
  \small
  \caption{\textbf{What the supervision requires.} The rows share the generator, budget and schedule and differ only in
  what the labels require, which is order-invariant, partly order-free or order-necessary, and the order-invariant
  labels are posed on the \ours clips. Change in accuracy (points) vs the base model is the mean \std{sd} over $n$
  seeds. Held-out is accuracy (\%) on the order-necessary curriculum test split, and Vino.\ and TimeB.\ are
  the text score and instance accuracy on Vinoground and TimeBlind.}
  \label{tab:crit}
  \setlength{\tabcolsep}{3pt}
  \adjustbox{max width=\linewidth}{\begin{tabular}{@{}lccccccc@{}}
    \toprule
    Labels & $n$ & TOMATO & w/o TOM. & Mean & Held-out & Vino. & TimeB. \\
    \midrule
    \multicolumn{8}{@{}l}{\emph{Qwen3.5-9B}} \\
Order-invariant & 2 & +1.3 & +2.1 & +1.9\std{0.3} & 52.8 & +4.5 & +6.6 \\
Leaky & 3 & +16.4 & +2.1 & +5.7\std{0.6} & 65.7 & +6.6 & +15.6 \\
\ours & 3 & +21.7 & +3.0 & \textbf{+7.6\std{0.3}} & 81.1 & +7.7 & +17.7 \\
\quad shuffled & 1 & +4.3 & $-$0.4 & +0.8 & 52.0 & $-$1.6 & $-$1.7 \\
\midrule
\multicolumn{8}{@{}l}{\emph{Qwen3-VL-8B}} \\
Order-invariant & 1 & $-$4.4 & $-$0.3 & $-$1.4 & 48.0 & +2.8 & $-$0.2 \\
Leaky & 2 & +13.1 & +1.4 & +4.4\std{0.3} & 64.6 & +5.1 & +12.1 \\
\ours & 3 & +13.8 & +2.0 & \textbf{+4.9\std{0.1}} & 76.7 & +5.6 & +13.1 \\
\quad shuffled & 1 & $-$0.6 & $-$0.8 & $-$0.8 & 47.5 & +1.6 & +2.3 \\
 
    \bottomrule
  \end{tabular}}
\end{table}

\paragraph{What the labels require.}
\label{sec:leak}
\cref{tab:crit} varies only what the labels require. For \emph{order-invariant labels} we keep the ordered clips and
ask questions that the unordered frames answer, such as the axis of motion, the position in the frame, which ball is
higher or how many balls there are. A \emph{leaky} curriculum ties static cues to the labels of speed change,
direction, trend and count, for example clip length to speed, so that an order-free classifier predicts translational
speed perfectly, while start order, the start and end counts of movers and the rotation, oscillation and approach clips
still require order. The gain grows with the share of order-necessary supervision, from \oiMean{} to \leakMeanThree{}
and \gainQnine{} points on Qwen3.5-9B and from \vloiMean{} to \vlleakMeanTwo{} and \gainQvl{} on Qwen3-VL-8B. Paired by
seed, \ours exceeds order-invariant labels on TOMATO by 18.7 and 21.4 points for Qwen3.5-9B and by 18.3 for Qwen3-VL-8B,
and it exceeds the leaky curriculum by 3.0 to 6.8 points on Qwen3.5-9B with 95\% intervals above zero and by 0.5 to 0.9
on Qwen3-VL-8B, as expected when part of the labels remain order-necessary. The two held-out
minimal-pair benchmarks give the same ranking of means. Paired by seed on TimeBlind, \ours leads its shuffled control by 18.8 to
19.8 and 9.8 to 12.0 points and order-invariant labels by 10.7 to 11.5 and 12.3 to 14.5 points. On Vinoground it leads
its shuffled control by 8.7 to 9.7 and 3.3 to 4.8 points and order-invariant labels by 2.1 to 3.6 points, across all
seeds of both backbones.

\paragraph{Curriculum components and real footage.}
Removing one primitive at a time shows that speed change is the only primitive with a large specific effect, costing
29.0 points on TOMATO velocity, while every variant keeps 78 to 95\% of the TVBench and TempCompass gain. Trained only on the real-footage constructions with
4{,}000 clips, Qwen3.5-9B gains +2.9 points on the four-benchmark mean, with +8.4 on TOMATO, and Qwen3-VL-8B gains +1.9,
against $-0.7$ for Qwen3.5-9B with shuffled frames. On held-out real windows the Qwen3.5-9B model reaches 70.0\%, against
37.3\% for the base model and 40.0\% for the shuffled control, so the criterion also transfers to unlabelled real footage.
Varying the LoRA rank, the learning rate or the number of clips keeps the four-benchmark mean between +6.7 and +8.4
points, and without replay MotionBench changes by $-8.8$ and MVBench by $-1.1$ points.

\subsection{Analysis}
\label{sec:analysis}

\begin{table}[t]
  \centering
  \small
  \caption{\textbf{Test-time frame order.} Accuracy (\%) with the test frames in their original order (Ord.),
  shuffled (Shuf.) and reversed (Rev.), on identical items, with TOMATO on the 1{,}385 questions with unique keys.
  Models are not retrained.}
  \label{tab:testorder}
  \setlength{\tabcolsep}{3.5pt}
  \adjustbox{max width=\linewidth}{\begin{tabular}{@{}lcccccc@{}}
    \toprule
    & \multicolumn{3}{c}{TOMATO} & \multicolumn{3}{c}{TVBench} \\
    \cmidrule(lr){2-4}\cmidrule(lr){5-7}
    & Ord. & Shuf. & Rev. & Ord. & Shuf. & Rev. \\
    \midrule
    \multicolumn{7}{l}{\textit{Qwen3.5-9B}} \\
\quad Base model & 32.1 & 23.3 & 22.3 & 56.1 & 37.6 & 28.9 \\
\quad \ours{} & 54.6 & 25.6 & 24.5 & 62.4 & 37.1 & 30.0 \\
\midrule
\multicolumn{7}{l}{\textit{Qwen3-VL-8B}} \\
\quad Base model & 33.6 & 24.9 & 24.6 & 53.9 & 41.2 & 31.9 \\
\quad \ours{} & 48.6 & 24.3 & 23.7 & 57.4 & 41.2 & 29.8 \\
 
    \bottomrule
  \end{tabular}}
\end{table}

\paragraph{Does the model use frame order at test time?}
\label{sec:testorder}
\cref{tab:testorder} evaluates the trained models with perturbed test frames. Shuffling or reversing the TOMATO test
frames lowers the accuracy of the base models by about 9 points and that of \ours by 24 to 30 points, to within 2.3
points of the base model under the same perturbation, and TVBench behaves likewise for shuffled frames. Compared with
fully shuffled frames, shuffling intact two-frame patches recovers 1.4 to 1.5 points for every model, so \ours uses
order across patches, in line with~\cite{rahman2026channel}. For the 457 TOMATO questions whose answer has a
time-reversal partner, \ours selects that partner on reversed clips, for example decelerating instead of
accelerating, in 54.5\% and 47.3\% of cases, against 19.9\% and 25.8\% for the base models.

\begin{figure}[t]
  \centering
  \includegraphics[width=\linewidth]{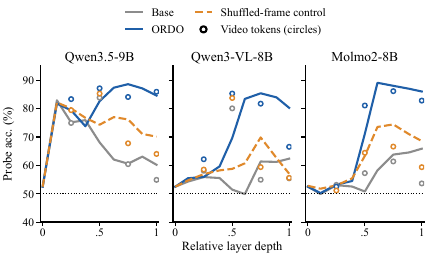}
  \caption{\textbf{Linear decodability of accelerating versus decelerating} on TOMATO across the depth of the language
  model (164 clips, grouped cross-validation). Lines show the last-token representation and circles the video tokens.
  Chance is 50\%.}
  \label{fig:probes}
\end{figure}

\paragraph{Where the motion signal forms.}
\label{sec:probe}
With the vision encoder frozen, temporal information must either be computed by the language model or already be
present, and layer-wise linear probes distinguish the two. We probe the 164 TOMATO clips from 83 source videos whose
answer is ``accelerating'' or ``decelerating''. At each probed layer we extract the hidden state at the answer position,
the \emph{last token}, and the video-token states pooled spatially at each time step and concatenated in temporal order,
the \emph{video tokens}. A logistic-regression probe on standardised features reduced to 32 principal components, both
fitted within each training fold, predicts the label, and we average its accuracy over three seeds under 5-fold
cross-validation grouped by source video~\cite{alain2016understanding,belinkov2022probing}. Probes on permuted labels
stay at 52 to 53\%, and the question text alone predicts the label at 53.6\% against a majority rate of 52.4\%, so the
probes read visual information. With 164 clips the 95\% interval is about $\pm 7$ points. \emph{Forced choice} compares the model's logits for the two options.
\cref{fig:probes} reveals two regimes. In base Qwen3.5-9B the label is decodable from the last token at layer 4 with
83\% and from mid-layer video tokens with 84\%, but last-token accuracy drops to 60\% at the final layer and the forced
choice is at chance with 52\%. Base Qwen3-VL-8B encodes it in mid-layer video tokens with 80\% while its last token stays
at or below 62\%. After training, the last token carries the label through the final layers with 85\% and 80\%, the
forced choice rises to 90\% and 84\%, and the video tokens change little, from 84\% to 87\% and from 80\% to 85\%, so
training makes information that the base model already computes available at the answer position. For base Molmo2-8B,
which unlike the Qwen models encodes every frame separately, pooled linear probes decode the label poorly, at most 66\% in the last token and 61\% in video tokens. After training its video tokens reach 86\% and its
last token 89\%, with a forced choice of 90\%, while the frozen input features stay at 53\%, so here training forms a
decodable representation. In both regimes the shuffled-frame controls stay below \ours in the final layers, with a
forced choice near chance at 54 to 57\%. Removing a decodable subspace of 1 to 16 directions changes the forced choice by
at most 1.2 points, close to random subspaces at 0.6, so the regimes show where the label becomes linearly available.

\section{Conclusion}
\label{sec:conclusion}

We presented \ours, a framework that makes frame order necessary for the training labels of a video LLM. It defines
order-necessary supervision as data whose unordered frames have the same distribution under competing labels, realises
it for motion attributes with time-reversal pairing, matched spatial extent and uniform interval coverage, and trains
low-rank adapters on a procedural curriculum with self-distillation replay. Across three
video LLMs and two scales, \ours raises the mean over four motion benchmarks by 4.9 to 7.6 points and improves held-out minimal pairs, the
shuffled-frame controls change the mean by less than one point, and the gain grows with the share of order-necessary labels. Probing reveals two regimes, in which training exposes a motion signal that the base model already computes or forms a
new linearly decodable one.

\section*{LLM Usage Statement}
For manuscript preparation, we used Claude (Anthropic) solely to correct grammar and improve the clarity and
readability of author-written text. All scientific contributions are the original work of the authors. The authors
reviewed all edits and take full responsibility for the manuscript.

\FloatBarrier
{\small\bibliographystyle{ieeenat_fullname}\bibliography{refs}}

\begin{thebibliography}{54}
\providecommand{\natexlab}[1]{#1}
\providecommand{\url}[1]{\texttt{#1}}
\expandafter\ifx\csname urlstyle\endcsname\relax
  \providecommand{\doi}[1]{doi: #1}\else
  \providecommand{\doi}{doi: \begingroup \urlstyle{rm}\Url}\fi

\bibitem[Alain and Bengio(2016)]{alain2016understanding}
Guillaume Alain and Yoshua Bengio.
\newblock Understanding intermediate layers using linear classifier probes.
\newblock \emph{arXiv preprint arXiv:1610.01644}, 2016.

\bibitem[Bagad et~al.(2023)Bagad, Tapaswi, and Snoek]{bagad2023test}
Piyush Bagad, Makarand Tapaswi, and Cees G.~M. Snoek.
\newblock Test of time: Instilling video-language models with a sense of time.
\newblock In \emph{Proceedings of the IEEE/CVF Conference on Computer Vision
  and Pattern Recognition (CVPR)}, pages 2503--2516, 2023.

\bibitem[Bai et~al.(2025)Bai, Cai, Chen, Chen, Chen, Cheng, Deng, Ding, Gao,
  Ge, Ge, Guo, Huang, Huang, Huang, Hui, Jiang, Li, Li, Li, Li, Lin, Lin, Liu,
  Liu, Liu, Liu, Liu, Liu, Lu, Luo, Lv, Men, Meng, Ren, Ren, Song, Sun, Tang,
  Tu, Wan, Wang, Wang, Wang, Wang, Xie, Xu, Xu, Xu, Yang, Yang, Yang, Yang, Yu,
  Zhang, Zhang, Zhang, Zheng, Zhong, Zhou, Zhou, Zhou, Zhu, and
  Zhu]{bai2025qwen3vl}
Shuai Bai, Yuxuan Cai, Ruizhe Chen, Keqin Chen, Xionghui Chen, Zesen Cheng,
  Lianghao Deng, Wei Ding, Chang Gao, Chunjiang Ge, Wenbin Ge, Zhifang Guo,
  Qidong Huang, Jie Huang, Fei Huang, Binyuan Hui, Shutong Jiang, Zhaohai Li,
  Mingsheng Li, Mei Li, Kaixin Li, Zicheng Lin, Junyang Lin, Xuejing Liu,
  Jiawei Liu, Chenglong Liu, Yang Liu, Dayiheng Liu, Shixuan Liu, Dunjie Lu,
  Ruilin Luo, Chenxu Lv, Rui Men, Lingchen Meng, Xuancheng Ren, Xingzhang Ren,
  Sibo Song, Yuchong Sun, Jun Tang, Jianhong Tu, Jianqiang Wan, Peng Wang,
  Pengfei Wang, Qiuyue Wang, Yuxuan Wang, Tianbao Xie, Yiheng Xu, Haiyang Xu,
  Jin Xu, Zhibo Yang, Mingkun Yang, Jianxin Yang, An Yang, Bowen Yu, Fei Zhang,
  Hang Zhang, Xi Zhang, Bo Zheng, Humen Zhong, Jingren Zhou, Fan Zhou, Jing
  Zhou, Yuanzhi Zhu, and Ke Zhu.
\newblock {Qwen3-VL} technical report.
\newblock \emph{arXiv preprint arXiv:2511.21631}, 2025.

\bibitem[Belinkov(2022)]{belinkov2022probing}
Yonatan Belinkov.
\newblock Probing classifiers: Promises, shortcomings, and advances.
\newblock \emph{Computational Linguistics}, 48\penalty0 (1):\penalty0 207--219,
  2022.

\bibitem[Buch et~al.(2022)Buch, Eyzaguirre, Gaidon, Wu, Fei-Fei, and
  Niebles]{buch2022revisiting}
Shyamal Buch, Crist{\'o}bal Eyzaguirre, Adrien Gaidon, Jiajun Wu, Li Fei-Fei,
  and Juan~Carlos Niebles.
\newblock Revisiting the ``video'' in video-language understanding.
\newblock In \emph{Proceedings of the IEEE/CVF Conference on Computer Vision
  and Pattern Recognition (CVPR)}, pages 2917--2927, 2022.

\bibitem[Clark et~al.(2019)Clark, Yatskar, and Zettlemoyer]{clark2019dont}
Christopher Clark, Mark Yatskar, and Luke Zettlemoyer.
\newblock Don't take the easy way out: Ensemble based methods for avoiding
  known dataset biases.
\newblock In \emph{Proceedings of the 2019 Conference on Empirical Methods in
  Natural Language Processing and the 9th International Joint Conference on
  Natural Language Processing (EMNLP-IJCNLP)}, pages 4069--4082, 2019.

\bibitem[Clark et~al.(2026)Clark, Zhang, Ma, Park, Tripathi, Lee, Salehi, Ren,
  Kim, Yang, Shao, Yang, Huang, Gao, Anderson, Zhang, Jain, Stoica, Farhadi,
  and Krishna]{clark2026molmo2}
Christopher Clark, Jieyu Zhang, Zixian Ma, Jae~Sung Park, Rohun Tripathi,
  Sangho Lee, Mohammadreza Salehi, Jason Ren, Chris~Dongjoo Kim, Yinuo Yang,
  Vincent Shao, Yue Yang, Weikai Huang, Ziqi Gao, Taira Anderson, Jianrui
  Zhang, Jitesh Jain, George Stoica, Ali Farhadi, and Ranjay Krishna.
\newblock {Molmo2}: Open weights and data for vision-language models with video
  understanding and grounding.
\newblock In \emph{Proceedings of the IEEE/CVF Conference on Computer Vision
  and Pattern Recognition (CVPR)}, pages 28652--28668, 2026.

\bibitem[Cores et~al.(2025)Cores, Dorkenwald, Mucientes, Snoek, and
  Asano]{cores2025lost}
Daniel Cores, Michael Dorkenwald, Manuel Mucientes, Cees G.~M. Snoek, and
  Yuki~M. Asano.
\newblock Lost in time: A new temporal benchmark for {VideoLLMs}.
\newblock In \emph{British Machine Vision Conference (BMVC)}, 2025.

\bibitem[Du et~al.(2026{\natexlab{a}})Du, Liu, Qin, Han, Gu, Zhu, Zhang, Liu,
  Chen, and Guo]{du2026crpo}
Dazhao Du, Jian Liu, Jialong Qin, Tao Han, Bohai Gu, Fangqi Zhu, Yujia Zhang,
  Eric Liu, Xi Chen, and Song Guo.
\newblock Learning spatiotemporal sensitivity in video {LLMs} via
  counterfactual reinforcement learning.
\newblock \emph{arXiv preprint arXiv:2605.21988}, 2026{\natexlab{a}}.

\bibitem[Du et~al.(2024)Du, Liu, and Jin]{du2024reversed}
Yang Du, Yuqi Liu, and Qin Jin.
\newblock Reversed in time: A novel temporal-emphasized benchmark for
  cross-modal video-text retrieval.
\newblock In \emph{Proceedings of the ACM International Conference on
  Multimedia (ACM MM)}, pages 5260--5269, 2024.

\bibitem[Du et~al.(2026{\natexlab{b}})Du, Fan, Nan, Xie, Zhou, Li, Yang, Yang,
  and Tai]{du2026motionsight}
Yipeng Du, Tiehan Fan, Kepan Nan, Rui Xie, Penghao Zhou, Xiang Li, Jian Yang,
  Zhenheng Yang, and Ying Tai.
\newblock {MotionSight}: Boosting fine-grained motion understanding in
  multimodal {LLMs}.
\newblock In \emph{International Conference on Learning Representations
  (ICLR)}, 2026{\natexlab{b}}.

\bibitem[Efron and Tibshirani(1993)]{efron1993introduction}
Bradley Efron and Robert~J. Tibshirani.
\newblock \emph{An Introduction to the Bootstrap}.
\newblock Number~57 in Monographs on Statistics and Applied Probability.
  Chapman \& Hall, New York, 1993.

\bibitem[Elazar et~al.(2021)Elazar, Ravfogel, Jacovi, and
  Goldberg]{elazar2021amnesic}
Yanai Elazar, Shauli Ravfogel, Alon Jacovi, and Yoav Goldberg.
\newblock Amnesic probing: Behavioral explanation with amnesic counterfactuals.
\newblock \emph{Transactions of the Association for Computational Linguistics},
  9:\penalty0 160--175, 2021.

\bibitem[Feng et~al.(2025)Feng, Gong, Li, Guo, Wang, Peng, Wu, Zhang, Wang, and
  Yue]{feng2025videor1}
Kaituo Feng, Kaixiong Gong, Bohao Li, Zonghao Guo, Yibing Wang, Tianshuo Peng,
  Junfei Wu, Xiaoying Zhang, Benyou Wang, and Xiangyu Yue.
\newblock {Video-R1}: Reinforcing video reasoning in {MLLMs}.
\newblock In \emph{Advances in Neural Information Processing Systems
  (NeurIPS)}, pages 99114--99137, 2025.

\bibitem[Fernando et~al.(2017)Fernando, Bilen, Gavves, and
  Gould]{fernando2017self}
Basura Fernando, Hakan Bilen, Efstratios Gavves, and Stephen Gould.
\newblock Self-supervised video representation learning with odd-one-out
  networks.
\newblock In \emph{Proceedings of the IEEE Conference on Computer Vision and
  Pattern Recognition (CVPR)}, pages 3636--3645, 2017.

\bibitem[Fu et~al.(2025)Fu, Dai, Luo, Li, Ren, Zhang, Wang, Zhou, Shen, Zhang,
  Chen, Li, Lin, Zhao, Li, Xu, Zheng, Chen, Shan, He, and Sun]{fu2025videomme}
Chaoyou Fu, Yuhan Dai, Yongdong Luo, Lei Li, Shuhuai Ren, Renrui Zhang, Zihan
  Wang, Chenyu Zhou, Yunhang Shen, Mengdan Zhang, Peixian Chen, Yanwei Li,
  Shaohui Lin, Sirui Zhao, Ke Li, Tong Xu, Xiawu Zheng, Enhong Chen, Caifeng
  Shan, Ran He, and Xing Sun.
\newblock {Video-MME}: The first-ever comprehensive evaluation benchmark of
  multi-modal {LLMs} in video analysis.
\newblock In \emph{Proceedings of the IEEE/CVF Conference on Computer Vision
  and Pattern Recognition (CVPR)}, pages 24108--24118, 2025.

\bibitem[Ghazanfari et~al.(2026)Ghazanfari, Garg, Krishnamurthy, and
  Khorrami]{ghazanfari2026syncr}
Sara Ghazanfari, Siddharth Garg, Prashanth Krishnamurthy, and Farshad Khorrami.
\newblock {SYNCR}: Diagnosing and learning cross-video reasoning from
  simulation.
\newblock \emph{arXiv preprint arXiv:2609.37918}, 2026.

\bibitem[Greff et~al.(2022)Greff, Belletti, Beyer, Doersch, Du, Duckworth,
  Fleet, Gnanapragasam, Golemo, Herrmann, Kipf, Kundu, Lagun, Laradji, Liu,
  Meyer, Miao, Nowrouzezahrai, Oztireli, Pot, Radwan, Rebain, Sabour, Sajjadi,
  Sela, Sitzmann, Stone, Sun, Vora, Wang, Wu, Yi, Zhong, and
  Tagliasacchi]{greff2022kubric}
Klaus Greff, Francois Belletti, Lucas Beyer, Carl Doersch, Yilun Du, Daniel
  Duckworth, David~J. Fleet, Dan Gnanapragasam, Florian Golemo, Charles
  Herrmann, Thomas Kipf, Abhijit Kundu, Dmitry Lagun, Issam Laradji,
  Hsueh-Ti~(Derek) Liu, Henning Meyer, Yishu Miao, Derek Nowrouzezahrai, Cengiz
  Oztireli, Etienne Pot, Noha Radwan, Daniel Rebain, Sara Sabour, Mehdi S.~M.
  Sajjadi, Matan Sela, Vincent Sitzmann, Austin Stone, Deqing Sun, Suhani Vora,
  Ziyu Wang, Tianhao Wu, Kwang~Moo Yi, Fangcheng Zhong, and Andrea
  Tagliasacchi.
\newblock {Kubric}: A scalable dataset generator.
\newblock In \emph{Proceedings of the IEEE/CVF Conference on Computer Vision
  and Pattern Recognition (CVPR)}, pages 3749--3761, 2022.

\bibitem[Han et~al.(2026)Han, Cheng, Pereira, Liu, and
  Kitazawa]{han2026tracing}
Peitao Han, Fei Cheng, Lis~K. Pereira, Qianying Liu, and Shigeru Kitazawa.
\newblock Tracing the arrow of time: Diagnosing temporal information flow in
  {Video-LLMs}.
\newblock \emph{arXiv preprint arXiv:2605.07568}, 2026.

\bibitem[He et~al.(2026)He, Qu, Li, Huang, Liu, and Cheng]{he2025videossr}
Zefeng He, Xiaoye Qu, Yafu Li, Siyuan Huang, Daizong Liu, and Yu Cheng.
\newblock {VideoSSR}: Video self-supervised reinforcement learning.
\newblock In \emph{Proceedings of the IEEE/CVF Conference on Computer Vision
  and Pattern Recognition (CVPR)}, pages 26530--26540, 2026.

\bibitem[Hinton et~al.(2015)Hinton, Vinyals, and Dean]{hinton2015distilling}
Geoffrey Hinton, Oriol Vinyals, and Jeff Dean.
\newblock Distilling the knowledge in a neural network.
\newblock \emph{arXiv preprint arXiv:1503.02531}, 2015.

\bibitem[Hong et~al.(2025)Hong, Cheng, Yang, Wang, Wang, Gu, Huang, Dong, and
  Tang]{hong2025motionbench}
Wenyi Hong, Yean Cheng, Zhuoyi Yang, Weihan Wang, Lefan Wang, Xiaotao Gu, Shiyu
  Huang, Yuxiao Dong, and Jie Tang.
\newblock {MotionBench}: Benchmarking and improving fine-grained video motion
  understanding for vision language models.
\newblock In \emph{Proceedings of the IEEE/CVF Conference on Computer Vision
  and Pattern Recognition (CVPR)}, pages 8450--8460, 2025.

\bibitem[Hu et~al.(2022)Hu, Shen, Wallis, Allen-Zhu, Li, Wang, Wang, and
  Chen]{hu2022lora}
Edward~J. Hu, Yelong Shen, Phillip Wallis, Zeyuan Allen-Zhu, Yuanzhi Li, Shean
  Wang, Lu Wang, and Weizhu Chen.
\newblock {LoRA}: Low-rank adaptation of large language models.
\newblock In \emph{International Conference on Learning Representations
  (ICLR)}, 2022.

\bibitem[Jiang et~al.(2026)Jiang, Song, Zhou, Wang, Chen, Guan, Luo, Zhang,
  Tang, Sun, Zhang, Yang, Bai, Lin, and Liu]{jiang2026learning}
Songtao Jiang, Sibo Song, Chenyi Zhou, Yuan Wang, Ruizhe Chen, Tongkun Guan,
  Ruilin Luo, Yan Zhang, Zhihang Tang, Yuchong Sun, Hang Zhang, Zhibo Yang,
  Shuai Bai, Junyang Lin, and Zuozhu Liu.
\newblock Learning transferable temporal primitives for video reasoning via
  synthetic videos.
\newblock In \emph{Proceedings of the IEEE/CVF Conference on Computer Vision
  and Pattern Recognition (CVPR)}, pages 31283--31292, 2026.

\bibitem[Khoshnoodi and Bargal(2026)]{khoshnoodi2026repprobe}
Mahsa Khoshnoodi and Sarah~Adel Bargal.
\newblock {RepProbe}: Can video language models count repeated actions? {A}
  diagnostic benchmark.
\newblock \emph{arXiv preprint arXiv:2610.18844}, 2026.

\bibitem[Kim et~al.(2026)Kim, Kim, and Han]{kim2026map}
Minji Kim, Taekyung Kim, and Bohyung Han.
\newblock Map the flow: Revealing hidden pathways of information in
  {VideoLLMs}.
\newblock In \emph{International Conference on Learning Representations
  (ICLR)}, 2026.

\bibitem[Kwon et~al.(2023)Kwon, Li, Zhuang, Sheng, Zheng, Yu, Gonzalez, Zhang,
  and Stoica]{kwon2023efficient}
Woosuk Kwon, Zhuohan Li, Siyuan Zhuang, Ying Sheng, Lianmin Zheng, Cody~Hao Yu,
  Joseph~E. Gonzalez, Hao Zhang, and Ion Stoica.
\newblock Efficient memory management for large language model serving with
  {PagedAttention}.
\newblock In \emph{Proceedings of the 29th Symposium on Operating Systems
  Principles (SOSP)}, pages 611--626. ACM, 2023.

\bibitem[Lee et~al.(2026)Lee, Lee, Kim, Kim, Chung, and Choi]{lee2026which}
Jongseo Lee, Hyuntak Lee, Sunghun Kim, Sooa Kim, Jihoon Chung, and Jinwoo Choi.
\newblock Which way did it move? {D}iagnosing and overcoming directional motion
  blindness in video {LLMs}.
\newblock In \emph{Advances in Neural Information Processing Systems
  (NeurIPS)}, 2026.

\bibitem[Lei et~al.(2023)Lei, Berg, and Bansal]{lei2023revealing}
Jie Lei, Tamara Berg, and Mohit Bansal.
\newblock Revealing single frame bias for video-and-language learning.
\newblock In \emph{Proceedings of the 61st Annual Meeting of the Association
  for Computational Linguistics (Volume 1: Long Papers)}, pages 487--507.
  Association for Computational Linguistics, 2023.

\bibitem[Li et~al.(2026{\natexlab{a}})Li, Zhao, Zhang, Mitra, Nyandwi, and
  Bertasius]{li2026timeblind}
Baiqi Li, Kangyi Zhao, Ce Zhang, Chancharik Mitra, Jean De~Dieu Nyandwi, and
  Gedas Bertasius.
\newblock {TimeBlind}: A spatio-temporal compositionality benchmark for video
  {LLMs}.
\newblock In \emph{Proceedings of the 2026 Conference on Empirical Methods in
  Natural Language Processing (EMNLP)}, pages 43801--43822, 2026{\natexlab{a}}.

\bibitem[Li et~al.(2024)Li, Wang, He, Li, Wang, Liu, Wang, Xu, Chen, Luo, Wang,
  and Qiao]{li2024mvbench}
Kunchang Li, Yali Wang, Yinan He, Yizhuo Li, Yi Wang, Yi Liu, Zun Wang, Jilan
  Xu, Guo Chen, Ping Luo, Limin Wang, and Yu Qiao.
\newblock {MVBench}: A comprehensive multi-modal video understanding benchmark.
\newblock In \emph{Proceedings of the IEEE/CVF Conference on Computer Vision
  and Pattern Recognition (CVPR)}, pages 22195--22206, 2024.

\bibitem[Li et~al.(2026{\natexlab{b}})Li, Li, Ouyang, Ren, Liu, Zhang, Zhang,
  Kong, Liu, and Sun]{li2026temple}
Shicheng Li, Lei Li, Kun Ouyang, Shuhuai Ren, Yuanxin Liu, Yuanxing Zhang,
  Fuzheng Zhang, Lingpeng Kong, Qi Liu, and Xu Sun.
\newblock {TEMPLE}: Incentivizing temporal understanding of video large
  language models via progressive pre-{SFT} alignment.
\newblock In \emph{Proceedings of the AAAI Conference on Artificial
  Intelligence (AAAI)}, pages 6378--6386, 2026{\natexlab{b}}.

\bibitem[Li and Hoiem(2018)]{li2018learning}
Zhizhong Li and Derek Hoiem.
\newblock Learning without forgetting.
\newblock \emph{IEEE Transactions on Pattern Analysis and Machine
  Intelligence}, 40\penalty0 (12):\penalty0 2935--2947, 2018.

\bibitem[Liu et~al.(2024)Liu, Li, Liu, Wang, Ren, Li, Chen, Sun, and
  Hou]{liu2024tempcompass}
Yuanxin Liu, Shicheng Li, Yi Liu, Yuxiang Wang, Shuhuai Ren, Lei Li, Sishuo
  Chen, Xu Sun, and Lu Hou.
\newblock {TempCompass}: Do video {LLMs} really understand videos?
\newblock In \emph{Findings of the Association for Computational Linguistics:
  {ACL} 2024}, pages 8731--8772. Association for Computational Linguistics,
  2024.

\bibitem[Liu et~al.(2026)Liu, Cao, Zheng, Liu, Qin, Qin, Wei, and
  Zhang]{liu2026relaxing}
Zijian Liu, Sihan Cao, Pengcheng Zheng, Kuien Liu, Caiyan Qin, Xiaolin Qin,
  Jiwei Wei, and Chaoning Zhang.
\newblock Relaxing anchor-frame dominance for mitigating hallucinations in
  video large language models.
\newblock \emph{arXiv preprint arXiv:2604.12582}, 2026.

\bibitem[Misra et~al.(2016)Misra, Zitnick, and Hebert]{misra2016shuffle}
Ishan Misra, C.~Lawrence Zitnick, and Martial Hebert.
\newblock Shuffle and learn: Unsupervised learning using temporal order
  verification.
\newblock In \emph{Proceedings of the European Conference on Computer Vision
  (ECCV)}, pages 527--544. Springer, 2016.

\bibitem[Pickup et~al.(2014)Pickup, Pan, Wei, Shih, Zhang, Zisserman,
  Sch{\"o}lkopf, and Freeman]{pickup2014seeing}
Lyndsey~C. Pickup, Zheng Pan, Donglai Wei, YiChang Shih, Changshui Zhang,
  Andrew Zisserman, Bernhard Sch{\"o}lkopf, and William~T. Freeman.
\newblock Seeing the arrow of time.
\newblock In \emph{Proceedings of the IEEE Conference on Computer Vision and
  Pattern Recognition (CVPR)}, 2014.

\bibitem[Price and Damen(2019)]{price2019retro}
Will Price and Dima Damen.
\newblock Retro-actions: Learning `close' by time-reversing `open' videos.
\newblock In \emph{Proceedings of the IEEE/CVF International Conference on
  Computer Vision Workshops (ICCVW)}, pages 1371--1380, 2019.

\bibitem[{Qwen Team}(2026)]{qwenteam2026qwen35}
{Qwen Team}.
\newblock {Qwen3.5}: Towards native multimodal agents.
\newblock \url{https://qwen.ai/blog?id=qwen3.5}, 2026.
\newblock Alibaba Cloud, blog post.

\bibitem[Rahman(2026)]{rahman2026channel}
Farrukh Rahman.
\newblock What does a temporal benchmark score measure? {D}ecomposing channel
  use in video {VLM} evaluation.
\newblock \emph{arXiv preprint arXiv:2607.12304}, 2026.

\bibitem[Ravfogel et~al.(2020)Ravfogel, Elazar, Gonen, Twiton, and
  Goldberg]{ravfogel2020null}
Shauli Ravfogel, Yanai Elazar, Hila Gonen, Michael Twiton, and Yoav Goldberg.
\newblock Null it out: Guarding protected attributes by iterative nullspace
  projection.
\newblock In \emph{Proceedings of the 58th Annual Meeting of the Association
  for Computational Linguistics (ACL)}, pages 7237--7256. Association for
  Computational Linguistics, 2020.

\bibitem[Sadhu and Cooper(2026)]{sadhu2026what}
Avyay Sadhu and Patrick Cooper.
\newblock What do verifiable rewards teach video-language models about time?
  {A} controlled multi-model study.
\newblock \emph{arXiv preprint arXiv:2610.03792}, 2026.

\bibitem[Shangguan et~al.(2025)Shangguan, Li, Ding, Zheng, Zhao, Fitzgerald,
  and Cohan]{shangguan2025tomato}
Ziyao Shangguan, Chuhan Li, Yuxuan Ding, Yanan Zheng, Yilun Zhao, Tesca
  Fitzgerald, and Arman Cohan.
\newblock {TOMATO}: Assessing visual temporal reasoning capabilities in
  multimodal foundation models.
\newblock In \emph{International Conference on Learning Representations
  (ICLR)}, 2025.

\bibitem[Shi et~al.(2026)Shi, Long, Wu, and Wang]{shi2026shape}
Yumeng Shi, Quanyu Long, Yin Wu, and Wenya Wang.
\newblock The shape of time: Video-token contrast for temporal understanding in
  {VideoLMs}.
\newblock In \emph{Proceedings of the 2026 Conference on Empirical Methods in
  Natural Language Processing (EMNLP)}, pages 36702--36717, 2026.

\bibitem[Shin et~al.(2026)Shin, Ro, Kim, and Kim]{shin2026before}
Youngwoo Shin, Yusung Ro, Minseo Kim, and Junmo Kim.
\newblock Before it fades: Reinforcing temporal representations at inference
  time in {VideoLLMs}.
\newblock In \emph{Advances in Neural Information Processing Systems
  (NeurIPS)}, 2026.

\bibitem[Sreeram et~al.(2026)Sreeram, Aasi, Wang, Maalouf, and
  Rus]{sreeram2026mind}
Shiva Sreeram, Erfan Aasi, Tsun-Hsuan Wang, Alaa Maalouf, and Daniela Rus.
\newblock Mind the temporal gap: Evaluating video models for temporal
  understanding in driving.
\newblock \emph{arXiv preprint arXiv:2610.12575}, 2026.

\bibitem[Vani et~al.(2025)Vani, Jena, Patel, Baral, Aditya, and
  Yang]{vani2025timewarp}
Sameep Vani, Shreyas Jena, Maitreya Patel, Chitta Baral, Somak Aditya, and
  Yezhou Yang.
\newblock Harnessing synthetic preference data for enhancing temporal
  understanding of {Video-LLMs}.
\newblock \emph{arXiv preprint arXiv:2510.03955}, 2025.

\bibitem[Wang et~al.(2025)Wang, He, Hong, Cheng, Zhang, Qi, Ding, Gu, Huang,
  Xu, Dong, and Tang]{wang2025lvbench}
Weihan Wang, Zehai He, Wenyi Hong, Yean Cheng, Xiaohan Zhang, Ji Qi, Ming Ding,
  Xiaotao Gu, Shiyu Huang, Bin Xu, Yuxiao Dong, and Jie Tang.
\newblock {LVBench}: An extreme long video understanding benchmark.
\newblock In \emph{Proceedings of the IEEE/CVF International Conference on
  Computer Vision (ICCV)}, pages 22958--22967, 2025.

\bibitem[Wei et~al.(2018)Wei, Lim, Zisserman, and Freeman]{wei2018learning}
Donglai Wei, Joseph~J. Lim, Andrew Zisserman, and William~T. Freeman.
\newblock Learning and using the arrow of time.
\newblock In \emph{Proceedings of the IEEE Conference on Computer Vision and
  Pattern Recognition (CVPR)}, pages 8052--8060, 2018.

\bibitem[Wu et~al.(2026)Wu, Zhang, Diao, Li, Lu, and Liu]{wu2025visualjigsaw}
Penghao Wu, Yushan Zhang, Haiwen Diao, Bo Li, Lewei Lu, and Ziwei Liu.
\newblock Visual jigsaw post-training improves {MLLMs}.
\newblock In \emph{International Conference on Learning Representations
  (ICLR)}, 2026.

\bibitem[Xu et~al.(2026)Xu, Qin, Jin, Lai, Cao, Huang, and Zhang]{xu2026tgpo}
Zhiyang Xu, Tian Qin, Bowen Jin, Zhengfeng Lai, Meng Cao, Lifu Huang, and Peng
  Zhang.
\newblock Incentivizing temporal-awareness in egocentric video understanding
  models.
\newblock \emph{arXiv preprint arXiv:2603.27184}, 2026.

\bibitem[Xue et~al.(2025)Xue, Luo, and Grauman]{xue2025seeing}
Zihui~(Sherry) Xue, Romy Luo, and Kristen Grauman.
\newblock Seeing the arrow of time in large multimodal models.
\newblock In \emph{Advances in Neural Information Processing Systems
  (NeurIPS)}, pages 90925--90955, 2025.

\bibitem[Yi et~al.(2020)Yi, Gan, Li, Kohli, Wu, Torralba, and
  Tenenbaum]{yi2020clevrer}
Kexin Yi, Chuang Gan, Yunzhu Li, Pushmeet Kohli, Jiajun Wu, Antonio Torralba,
  and Joshua~B. Tenenbaum.
\newblock {CLEVRER}: {CoLlision} events for video {REpresentation} and
  reasoning.
\newblock In \emph{International Conference on Learning Representations
  (ICLR)}, 2020.

\bibitem[Zhang et~al.(2024)Zhang, Cai, and Lee]{zhang2024vinoground}
Jianrui Zhang, Mu Cai, and Yong~Jae Lee.
\newblock {Vinoground}: Scrutinizing {LMMs} over dense temporal reasoning with
  short videos.
\newblock \emph{arXiv preprint arXiv:2410.02763}, 2024.

\end{thebibliography}
\clearpage
\appendix
\twocolumn[\centering\Large\textbf{Appendix}\vspace{1.0em}]
\setcounter{section}{0}\setcounter{table}{0}\setcounter{figure}{0}
\renewcommand{\thesection}{S\arabic{section}}
\renewcommand{\thetable}{S\arabic{table}}
\renewcommand{\thefigure}{S\arabic{figure}}
\section{Per-benchmark results and intervals}
\label{supp:ci}

\begin{table*}[t]
  \centering
  \small
  \caption{\textbf{Change in accuracy (points) with respect to the base model}, per benchmark and training run, with
  95\% paired-bootstrap intervals over items (2{,}000 resamples). Base accuracies (\%) for the five benchmarks are given after each
  backbone. s$k$ denotes the training seed, rows without s$k$ use seed 0, and MVBench was evaluated for seed 0 of \ours and for
  every control run.}
  \label{tab:supp_ci}
  \setlength{\tabcolsep}{4pt}
  \begin{tabular}{@{}lccccc@{}}
    \toprule
    & TOMATO & TVBench & MotionBench & TempCompass & MVBench \\
    \midrule
    \multicolumn{6}{@{}l}{\emph{Qwen3.5-9B} (base: 32.5, 56.1, 62.9, 75.1, 68.5)} \\
\ours s0 & +20.9\,{\scriptsize[+17.7, +23.9]} & +6.3\,{\scriptsize[+4.3, +8.3]} & $-$1.9\,{\scriptsize[$-$3.6, $-$0.3]} & +6.2\,{\scriptsize[+4.3, +8.2]} & +2.1\,{\scriptsize[+0.9, +3.3]} \\
\ours s1 & +21.9\,{\scriptsize[+18.9, +24.9]} & +5.8\,{\scriptsize[+3.9, +7.8]} & $-$2.0\,{\scriptsize[$-$3.6, $-$0.4]} & +5.3\,{\scriptsize[+3.4, +7.1]} & -- \\
\ours s2 & +22.3\,{\scriptsize[+19.3, +25.4]} & +4.8\,{\scriptsize[+3.0, +6.8]} & $-$3.2\,{\scriptsize[$-$5.0, $-$1.4]} & +5.4\,{\scriptsize[+3.4, +7.5]} & -- \\
Shuffled & +4.3\,{\scriptsize[+1.6, +6.9]} & $-$1.1\,{\scriptsize[$-$3.1, +1.0]} & $-$1.0\,{\scriptsize[$-$2.3, +0.4]} & +0.8\,{\scriptsize[$-$0.9, +2.5]} & +0.7\,{\scriptsize[$-$0.4, +1.8]} \\
Random label & $-$7.5\,{\scriptsize[$-$10.6, $-$4.6]} & $-$7.3\,{\scriptsize[$-$9.3, $-$5.4]} & $-$1.8\,{\scriptsize[$-$3.3, $-$0.3]} & $-$4.6\,{\scriptsize[$-$6.5, $-$2.7]} & $-$3.9\,{\scriptsize[$-$5.1, $-$2.7]} \\
Order-inv. s0 & +2.2\,{\scriptsize[$-$0.1, +4.5]} & +3.4\,{\scriptsize[+1.5, +5.5]} & $-$0.3\,{\scriptsize[$-$1.8, +1.3]} & +3.0\,{\scriptsize[+1.3, +4.7]} & +1.0\,{\scriptsize[$-$0.1, +2.2]} \\
Order-inv. s1 & +0.5\,{\scriptsize[$-$1.9, +2.8]} & +4.5\,{\scriptsize[+2.7, +6.4]} & $-$0.6\,{\scriptsize[$-$2.2, +0.9]} & +2.1\,{\scriptsize[+0.4, +3.7]} & +1.7\,{\scriptsize[+0.6, +2.9]} \\
Leaky s0 & +17.9\,{\scriptsize[+15.0, +20.8]} & +5.5\,{\scriptsize[+3.6, +7.6]} & $-$1.7\,{\scriptsize[$-$3.5, 0.0]} & +3.6\,{\scriptsize[+1.8, +5.5]} & +2.3\,{\scriptsize[+1.2, +3.6]} \\
Leaky s1 & +15.1\,{\scriptsize[+12.2, +17.9]} & +4.2\,{\scriptsize[+2.2, +6.3]} & $-$1.8\,{\scriptsize[$-$3.5, $-$0.2]} & +4.1\,{\scriptsize[+2.2, +5.9]} & +1.2\,{\scriptsize[$-$0.1, +2.4]} \\
Leaky s2 & +16.0\,{\scriptsize[+13.2, +18.9]} & +4.0\,{\scriptsize[+2.0, +6.1]} & $-$1.4\,{\scriptsize[$-$3.0, +0.3]} & +2.7\,{\scriptsize[+0.9, +4.6]} & +1.1\,{\scriptsize[0.0, +2.3]} \\
\midrule
\multicolumn{6}{@{}l}{\emph{Qwen3-VL-8B} (base: 34.4, 53.9, 57.0, 73.2, 68.1)} \\
\ours s0 & +13.8\,{\scriptsize[+11.1, +16.6]} & +3.5\,{\scriptsize[+1.6, +5.3]} & $-$1.2\,{\scriptsize[$-$2.9, +0.4]} & +3.0\,{\scriptsize[+1.3, +4.9]} & +0.4\,{\scriptsize[$-$0.6, +1.5]} \\
\ours s1 & +13.7\,{\scriptsize[+11.0, +16.6]} & +4.4\,{\scriptsize[+2.6, +6.2]} & $-$1.6\,{\scriptsize[$-$3.1, $-$0.1]} & +3.1\,{\scriptsize[+1.5, +4.9]} & -- \\
\ours s2 & +13.8\,{\scriptsize[+11.1, +16.6]} & +4.6\,{\scriptsize[+2.8, +6.3]} & $-$1.2\,{\scriptsize[$-$2.8, +0.4]} & +2.9\,{\scriptsize[+1.3, +4.6]} & -- \\
Shuffled & $-$0.6\,{\scriptsize[$-$3.2, +2.0]} & $-$2.4\,{\scriptsize[$-$4.3, $-$0.7]} & $-$0.2\,{\scriptsize[$-$1.5, +1.1]} & +0.1\,{\scriptsize[$-$1.7, +1.9]} & $-$0.1\,{\scriptsize[$-$1.1, +0.9]} \\
Order-inv. & $-$4.4\,{\scriptsize[$-$6.3, $-$2.6]} & $-$2.4\,{\scriptsize[$-$3.8, $-$1.0]} & 0.0\,{\scriptsize[$-$1.1, +1.2]} & +1.3\,{\scriptsize[0.0, +2.7]} & +0.3\,{\scriptsize[$-$0.6, +1.2]} \\
Leaky s0 & +13.3\,{\scriptsize[+10.6, +16.2]} & +3.7\,{\scriptsize[+1.9, +5.4]} & $-$0.3\,{\scriptsize[$-$1.9, +1.4]} & +1.7\,{\scriptsize[$-$0.1, +3.5]} & +2.1\,{\scriptsize[+1.0, +3.2]} \\
Leaky s1 & +12.9\,{\scriptsize[+10.1, +15.9]} & +4.2\,{\scriptsize[+2.2, +6.2]} & $-$2.2\,{\scriptsize[$-$4.0, $-$0.4]} & +1.6\,{\scriptsize[$-$0.3, +3.5]} & +1.2\,{\scriptsize[+0.1, +2.4]} \\
\midrule
\multicolumn{6}{@{}l}{\emph{Molmo2-8B} (base: 36.4, 56.9, 60.7, 72.1, 74.5)} \\
\ours s0 & +16.7\,{\scriptsize[+14.0, +19.4]} & +2.3\,{\scriptsize[+0.5, +4.0]} & $-$0.5\,{\scriptsize[$-$2.1, +1.2]} & +1.3\,{\scriptsize[$-$0.5, +3.1]} & $-$1.4\,{\scriptsize[$-$2.4, $-$0.5]} \\
\ours s1 & +17.9\,{\scriptsize[+15.2, +20.7]} & +3.9\,{\scriptsize[+2.3, +5.5]} & $-$0.6\,{\scriptsize[$-$2.0, +1.0]} & $-$0.1\,{\scriptsize[$-$2.0, +1.9]} & -- \\
Shuffled & +4.4\,{\scriptsize[+1.7, +7.1]} & $-$1.0\,{\scriptsize[$-$2.7, +0.7]} & $-$2.3\,{\scriptsize[$-$3.8, $-$0.7]} & $-$0.4\,{\scriptsize[$-$2.0, +1.3]} & $-$1.1\,{\scriptsize[$-$2.0, $-$0.2]} \\
\midrule
\multicolumn{6}{@{}l}{\emph{Qwen3.5-4B} (base: 33.9, 51.6, 59.3, 73.5, 65.5)} \\
\ours & +19.4\,{\scriptsize[+16.2, +22.4]} & +6.9\,{\scriptsize[+4.8, +9.0]} & $-$3.1\,{\scriptsize[$-$5.2, $-$1.0]} & +4.7\,{\scriptsize[+2.5, +6.8]} & +2.6\,{\scriptsize[+1.3, +3.9]} \\
Shuffled & +1.1\,{\scriptsize[$-$1.7, +4.0]} & $-$0.4\,{\scriptsize[$-$2.5, +1.7]} & +0.4\,{\scriptsize[$-$1.2, +2.0]} & +1.3\,{\scriptsize[$-$0.6, +3.3]} & +1.2\,{\scriptsize[0.0, +2.3]} \\
 
    \bottomrule
  \end{tabular}
\end{table*}

\cref{tab:supp_ci} lists every training run of \cref{tab:main,tab:crit} with its 95\% paired-bootstrap intervals.

\begin{table}[t]
  \centering
  \small
  \caption{\textbf{Four-benchmark mean with 95\% intervals.} Top: change in the mean over TOMATO, TVBench, MotionBench and
  TempCompass with respect to the base model, averaged over $n$ seeds, for \ours and its shuffled-frame control. Bottom:
  \ours (seed 0) minus each alternative training source of \cref{tab:compare}, paired on items. Each benchmark is resampled over
  its own items, with 2{,}000 resamples.}
  \label{tab:mean4}
  \setlength{\tabcolsep}{4pt}
  \begin{tabular}{@{}lccc@{}}
    \toprule
    & $n$ & \ours & Shuffled \\
    \midrule
    Qwen3.5-9B & 3 & +7.6\,{\scriptsize[+6.6, +8.7]} & +0.8\,{\scriptsize[$-$0.3, +1.7]} \\
Qwen3-VL-8B & 3 & +4.9\,{\scriptsize[+3.9, +5.8]} & $-$0.8\,{\scriptsize[$-$1.8, +0.1]} \\
Molmo2-8B & 2 & +5.1\,{\scriptsize[+4.2, +6.1]} & +0.2\,{\scriptsize[$-$0.7, +1.2]} \\
Qwen3.5-4B & 1 & +7.0\,{\scriptsize[+5.8, +8.2]} & +0.6\,{\scriptsize[$-$0.5, +1.6]} \\
 
    \midrule
    \ours minus & & Qwen3.5-9B & Qwen3-VL-8B \\
    \midrule
    SynRL data & & +5.5\,{\scriptsize[+4.5, +6.5]} & +3.9\,{\scriptsize[+2.9, +4.8]} \\
MoDirect-Inst & & +5.6\,{\scriptsize[+4.6, +6.6]} & -- \\
Temporal jigsaw, s0 & & +5.2\,{\scriptsize[+4.3, +6.2]} & +2.8\,{\scriptsize[+1.9, +3.7]} \\
Temporal jigsaw, s1 & & +4.7\,{\scriptsize[+3.7, +5.6]} & +1.2\,{\scriptsize[+0.3, +2.1]} \\
Arrow of time & & +7.8\,{\scriptsize[+6.8, +8.8]} & +4.6\,{\scriptsize[+3.7, +5.6]} \\
 
    \bottomrule
  \end{tabular}
\end{table}

\cref{tab:mean4} adds intervals for the four-benchmark mean. Every \ours gain excludes zero and every shuffled-frame
control includes it, and \ours exceeds each alternative training source of \cref{tab:compare} with an interval above zero on both
backbones it was compared on.

\section{Curriculum generator}
\label{supp:gen}

All clips are rendered procedurally and deterministically from (split, index). The training, validation and test splits use disjoint seeds, background stills, and question and option templates. Video-MME source videos are split 75/12.5/12.5 by video. Each question type has two to five training templates and one to three held-out templates. Frame sizes are drawn from
$448{\times}448$, $252{\times}448$, $448{\times}252$ and $336{\times}448$; clips last 4--12\,s (30--32\,s for counting) at 2
frames per second, and each frame is taken at the centre of its sampling interval, never at a clip boundary.

\paragraph{Speed change.} Progress along a path follows $p(u) = (u + \tfrac{a}{2} u^2)/(1 + \tfrac{a}{2})$ for speeding up
($a = k - 1$, where $k \in [2.5, 7]$ is the ratio of final to initial speed), its time reversal $1 - p(1-u)$ for slowing down and $p(u) = u$ for constant speed. The travel direction is drawn uniformly, so the three labels share the distribution of start point, end point and path. The same profile drives translation (balls, polygons or textured
discs, optionally on a curved path), rotation (at most 55$^\circ$ per frame so that the spin is not aliased at 2 frames per
second), camera pans over a still and the phase of an oscillation (frequencies of 0.12--0.45\,Hz, well below the Nyquist limit of 1\,Hz at 2 frames per second).
\paragraph{Direction and trend.} For the opposite label, translation and camera pans start at the other end of the same path, and approach and recession
change the sprite's scale exponentially between the same two sizes. Out-and-back trends cover the same interval at
constant speed as one-way motion, so both have the same position distribution.
\paragraph{Counting.} With $N$ drawn uniformly from $\{1, \dots, 12\}$, a clip shows $N$ cycles of a triangle wave with cycle lengths jittered by $\pm$15\% and
short still segments of at most 6\% at either end. The options are $N$ and three to five neighbours with the rank of $N$ drawn uniformly. A text-only classifier on the option values predicts the rank of the answer with 0.30 accuracy against 0.21 for chance, a residual from options near zero, where fewer lower neighbours exist.
\paragraph{Start order and number of movers.} In start order, both objects move once for the same duration, 30--42\% of the clip, and the second starts after a delay of 18--40\% of the clip duration. Lanes and directions are drawn independently of the label. In number of movers, early
and late movers move during the first 35--50\% or last 35--50\% of the clip, and questions about the start and the end are
exchanged by time reversal, and the option set is fixed, $0$--$4$ for movers throughout the clip and $0$--$3$ for start and end questions.
\paragraph{Distractors.} In 30\% of the speed, direction, trend and count clips the target is a named coloured ball among
one to three other balls moving at constant speed, and in 30\% of start-order clips one or two static balls are added.

\paragraph{Order-invariant and leaky variants.} These variants underlie \cref{tab:crit}. The order-invariant variant renders the identical
clips (byte-identical frames) and replaces each question with one about the axis of motion (horizontal, vertical, two
diagonals, or none), which third of the frame an object occupies (rotation, approach), which of the two moving balls is
higher (start order) or the total number of balls (number of movers), using existing generator state without further random draws. The leaky variant keeps every random draw but adds label-dependent static cues: label-specific clip length and a
fixed travel direction for speed, motion starting at the frame centre for direction and trend, a clip length that grows
with $N$ for counting, and smaller cues in the other primitives (a label-dependent size range for approach and
rotation, both objects moving rightwards in start order, early movers resting on one side). An order-free gradient-boosting classifier on permutation-invariant statistics of the moving object's position and size, together with the number of frames, then achieves an accuracy of 1.00 for translational speed (0.40 without the leak, majority 0.37), 0.95 for translational direction (0.52, majority 0.32), 0.80 for trend (0.27, majority 0.16) and 0.77 for count (0.07, majority 0.12). Without the leak, direction and trend are recovered only up to the axis of motion, which the unordered frames show. Within an axis, 0.52 and 0.27 match guessing between the reversal partners, which is the pairwise condition of \cref{def:on}. For start order, the start and end counts of movers, approach, rotation and oscillation the classifier stays within 0.05 of
the majority rate, so these labels show no order-free cue. Validation and test
clips are always rendered without the leak, so the two variants and \ours are all evaluated on identical held-out clips.

\section{Training and evaluation protocol}
\label{supp:protocol}

LoRA rank 16 ($\alpha = 32$, dropout 0.05) on the attention (and, in Qwen3.5, linear-attention) and MLP projections of
the language model; vision encoder and projector frozen; bf16; gradient checkpointing; 4{,}000 clips, 8 clips per AdamW step (500 steps), learning rate
$10^{-4}$ with 5\% warm-up and cosine decay, gradient clipping at 1.0. The loss is the cross-entropy of the answer letter
and the end-of-turn token. Self-distillation replay: we draw one replay item after each training clip (multiple-choice questions on Video-MME and LVBench videos, at most 32 frames, no gold labels used). We add the cross-entropy between the base model's and trained model's answer distributions over option letters with weight 1, which differs from the KL divergence of \cref{sec:objective} only by the constant entropy of the target, and 10\% of replay items are held out to measure agreement with the base model, which stays between 87\% and 99\% after training across runs. Clips whose processing would exceed available GPU memory are skipped; at
most 2 of 4{,}000 were skipped in any reported run. Evaluation uses vLLM with greedy decoding, uniform sampling at 2 frames per second up to 64 frames, a longer side of at most 448 pixels and a fixed instruction to answer with the option letter. We parse answers from the first option letter in the output, and an output without one counts as wrong. We key TOMATO items by (id, occurrence), since 36 ids are reused across 99 distinct questions. The reference implementation of TAI keys items by id, so \cref{tab:compare}, \cref{tab:testorder} and the frame-pair shuffle below drop these 99 items and use the remaining 1{,}385. MotionBench: a fixed random subset of 2{,}000 of the 4{,}018 development questions with public answers (the benchmark paper lists
4{,}020 development questions).
TVBench: nine tasks (the Action Antonym videos were not available). TempCompass: all 1{,}580 multiple-choice questions over its five temporal aspects. MVBench: 19 tasks, 3{,}800 questions (the
Fine-grained Pose videos require a separate licence).

\paragraph{Frame-pair shuffle.} Qwen3.5 and Qwen3-VL merge two consecutive frames into one temporal patch. Permuting
the order of the patches while keeping each pair intact gives TOMATO accuracies (\%, the 1{,}385 items of
\cref{tab:testorder}; ordered / pair-shuffled / shuffled) of 32.1 / 24.8 / 23.3 (Qwen3.5-9B), 54.6 / 27.1 / 25.6
(\ours), 33.6 / 26.4 / 24.9 (Qwen3-VL-8B) and 48.6 / 25.8 / 24.3 (\ours). Pair-shuffle minus shuffle is +1.4 to +1.5
points for all four models, with 95\% paired-bootstrap intervals from $[-1.2, +3.9]$ to $[-0.1, +3.2]$, so keeping
consecutive frames paired recovers little of the ordered accuracy.

\paragraph{Debiasing existing data instead of constructing it.} Order-necessary supervision could in principle be approximated by down-weighting what an order-blind model already predicts. We trained Qwen3.5-9B on SynRL with a product-of-experts objective~\cite{clark2019dont}. For each clip, we average the answer probabilities of the frozen Qwen3.5-9B model trained on shuffled SynRL frames over four frame permutations, take the element-wise logarithm and add the result to the model's logits during training and remove this correction at test time. The order-blind expert gives the correct SynRL answer a mean probability of 0.82 (above 0.9 for 61\% of
clips). The mean change is +2.2 points against +2.4 for plain SynRL training (TOMATO +5.7 against +7.3), and the drop
under test-time shuffling is similar (TOMATO 12.2 against 14.4 points, TVBench 20.2 against 18.3). In this setting, removing the order-free part of the signal does not create an order-dependent one, which supports building order necessity into the labels.

\paragraph{Baselines.} SynRL clips come from the official generators at their default sizes (500 per type for seven
atomic motion types, 200 per type for five extended types), subsampled to 4{,}000, and their question, options and answer
letter are used as given. MoDirect-Inst is subsampled to 4{,}000. TAI is run in its reference implementation against its own baseline, with $\beta = 0.5$ for Qwen3-VL-8B and $\beta = 1$ for
Qwen3.5-9B and Molmo2-8B. At the original 16 frames Qwen3-VL-8B gains +0.7 (TVBench, $\beta = 0.5$)
and +0.4 (TempCompass), with all intervals including zero. An order-free classifier on permutation-invariant per-pixel statistics of
SynRL clips predicts acceleration with 0.81 accuracy (majority 0.37), complex direction with 0.71 (0.11) and relative
direction with 1.00 (0.40), whereas for rotation count (0.33, majority 0.35) and rotation direction (0.53, majority 0.54) it stays at the majority rate.

\paragraph{Compute.} A 9B training run takes 3.8\,h on two RTX 3090 GPUs, and we also used RTX 4090 GPUs with 48\,GB each (a memory-upgraded variant). We evaluate every run on the four motion benchmarks and held-out curriculum split, and seed 0 of \ours and every control also on MVBench, under the same vLLM protocol throughout.

\section{Held-out minimal-pair benchmarks}
\label{supp:heldout}

Vinoground~\cite{zhang2024vinoground} and TimeBlind~\cite{li2026timeblind} were not used for any design decision, and their
protocol was fixed before any result was seen. We compute 95\% paired-bootstrap confidence intervals by resampling caption pairs for Vinoground and video pairs for TimeBlind. Comparisons are paired by seed. Single-seed controls are compared with every \ours seed, and an \ours seed without a
matching control seed is not compared with that control.

\paragraph{Vinoground.} Vinoground pairs natural videos whose captions use the same words in a different order (500
caption pairs; text score, two options, 1{,}000 questions). Over three seeds, \ours raises the text score of Qwen3.5-9B
from 76.2\% to 83.9\% (seeds +7.9, +8.1, +7.1; seed 0 [+5.6, +10.1]) and of Qwen3-VL-8B from 67.6\% to 73.2\% (+6.4, +5.5,
+4.9), and the pair score (both videos of a pair correct) rises by 13.8--15.8 and 9.6--11.4 points. Paired by seed, \ours leads its shuffled
control by 8.7 to 9.7 (Qwen3.5-9B) and 3.3 to 4.8 points (Qwen3-VL-8B), order-invariant labels by 3.4 to 3.5 and 2.1 to
3.6 (lowest interval [0.0, 4.1]), and the leaky curriculum by 0.6 to 1.5 and 0.6 to 1.2 (all intervals include zero).
All models are near chance (51.0--54.2\%) with shuffled test frames, so the gain depends on the order in which the frames are shown.

\paragraph{TimeBlind.} TimeBlind has 600 instances (2{,}400 video--question items, posed with our letter protocol). Each
instance is a minimal pair of near-identical videos (about 18\% simulated) with two complementary questions whose answers
flip between the videos. Of the instances, 300 use yes/no questions and 300 use two-option questions. The two halves reuse the same videos under
different file names, and by file content the 600 instances cover 280 distinct video pairs, which we resample as units. We report the official nested scores, in which an instance counts only if both
questions are answered correctly on both videos (I-Acc, chance 6.25\%). Over three seeds \ours raises I-Acc from 25.0\%
to 42.2--43.2\% (Qwen3.5-9B) and from 20.2\% to 32.3--34.5\% (Qwen3-VL-8B) and item accuracy from 66.5\% to 74.0\% and
from 63.5\% to 69.7--70.8\%. Paired by seed, I-Acc exceeds the shuffled control by 18.8 to 19.8 and 9.8 to 12.0 points,
order-invariant labels by 10.7 to 11.5 and 12.3 to 14.5 (all intervals above zero), and the leaky curriculum by 1.7 to
2.7 (intervals from $[-1.5, +4.8]$ to $[-1.0, +6.2]$) and by $-$0.8 and 3.2 (intervals $[-3.8, +2.2]$ and $[0.0, +6.4]$), so the
margin over the leaky curriculum is positive in four of five seed pairs but within the intervals. With shuffled test frames I-Acc falls to 6.7\% and 4.2\% for the base models and to 10.5\% and
7.0\% for \ours, against a chance level of 6.25\%.

\begin{table*}[t]
  \centering
  \small
  \caption{\textbf{Hyper-parameters and replay} (Qwen3.5-9B, seed 0). Each row changes one setting of the paper
  configuration. Columns give the change in accuracy (points) with respect to the base model on identical items, Mean is
  over the four motion benchmarks, and Held-out is accuracy (\%) on the held-out curriculum test split, where the base
  model reaches \mevalBase\%. The seed standard deviation of the paper configuration is 0.3 points for the mean.}
  \label{tab:hparam}
  \setlength{\tabcolsep}{5pt}
  \adjustbox{max width=\linewidth}{\begin{tabular}{@{}lccccccc@{}}
    \toprule
    & TOMATO & TVBench & MotionBench & TempCompass & Mean & MVBench & Held-out \\
    \midrule
    Paper setting (rank 16, lr $10^{-4}$, 4{,}000 clips, replay for every example) & \textbf{+20.9} & \textbf{+6.3} & \textbf{$-$1.9} & \textbf{+6.2} & \textbf{+7.9} & \textbf{+2.1} & \textbf{81.3} \\
\midrule
No replay & +21.0 & +5.6 & $-$8.8 & +6.4 & +6.1 & $-$1.1 & 79.8 \\
Replay for half of the examples & +21.9 & +6.2 & $-$2.5 & +7.2 & +8.2 & +1.8 & 81.2 \\
\midrule
LoRA rank 8 & +20.9 & +5.4 & $-$1.5 & +5.9 & +7.7 & +2.6 & 76.8 \\
LoRA rank 32 & +21.0 & +5.7 & $-$2.5 & +5.0 & +7.3 & +1.8 & 82.0 \\
\midrule
Learning rate $5{\times}10^{-5}$ & +17.9 & +5.6 & $-$1.2 & +5.7 & +7.0 & +2.1 & 76.7 \\
Learning rate $2{\times}10^{-4}$ & +20.2 & +4.7 & $-$3.9 & +5.6 & +6.7 & +1.2 & 83.8 \\
\midrule
2{,}000 clips (250 steps) & +20.6 & +5.6 & $-$1.8 & +5.3 & +7.4 & +2.3 & 73.0 \\
8{,}000 clips (1{,}000 steps) & +22.6 & +6.4 & $-$1.7 & +6.3 & +8.4 & +1.8 & 84.2 \\
 
    \bottomrule
  \end{tabular}}
\end{table*}

\section{Leave-one-primitive-out}
\label{sec:ablation}

\begin{table}[t]
  \centering
  \small
  \caption{\textbf{Leave-one-primitive-out} (Qwen3.5-9B, seed 0). Mean: four-benchmark mean change (points, two decimals because several variants differ by less than 0.1; the seed
  standard deviation of the full curriculum is 0.3). TOMATO velocity (its velocity and frequency category), TOMATO direction and MotionBench repetition count:
  change on those categories. TVB+TC: share of the full curriculum's TVBench and TempCompass gain that is retained (\%).
  Held-out: accuracy on the held-out curriculum test split (\%; base model \mevalBase). $^\ast$Distractor objects are a
  component of five primitives rather than a primitive. Leaky: the same generator with static cues to the label
  (\cref{sec:leak}).}
  \label{tab:ablation}
  \setlength{\tabcolsep}{3pt}
  \adjustbox{max width=\linewidth}{\begin{tabular}{@{}lcccccc@{}}
    \toprule
    & Mean & TOM.\ vel. & TOM.\ dir. & MoB.\ rep. & TVB+TC & Held-out \\
    \midrule
    Full curriculum & +7.87 & +53.3 & +34.5 & $-$8.7 & 100 & 81.3 \\
\midrule
$-$ Speed change & +6.89 & +24.3 & +36.5 & $-$8.7 & 85 & 67.8 \\
$-$ Direction & +7.64 & +53.3 & +27.8 & $-$9.2 & 95 & 78.2 \\
$-$ Out-and-back trend & +6.81 & +51.0 & +26.6 & $-$10.2 & 81 & 76.7 \\
$-$ Oscillation count & +7.55 & +52.9 & +29.0 & $-$6.3 & 92 & 78.8 \\
$-$ Start order & +7.08 & +54.3 & +25.6 & $-$8.7 & 88 & 79.3 \\
$-$ Number of movers & +6.93 & +49.5 & +30.0 & $-$10.2 & 78 & 79.5 \\
$-$ Distractors$^\ast$ & +7.77 & +53.3 & +35.0 & $-$10.7 & 88 & 75.5 \\
\midrule
Leaky curriculum & +6.34 & +39.0 & +31.0 & $-$9.7 & 73 & 66.7 \\
 
    \bottomrule
  \end{tabular}}
\end{table}

We remove each primitive in turn and reallocate its mixture share to the others, with results in \cref{tab:ablation}. Removing speed change reduces accuracy on TOMATO velocity questions by 29.0 points, and no other primitive has a comparably large category-specific effect. Direction is learned redundantly: accuracy on TOMATO direction falls by 6.7 points when the direction primitive is removed, and removing trend, start-order or count primitives, each also moving objects in a specific direction, has a similar effect. Removing speed, trend,
start order or number of movers lowers the mean by 0.8 to 1.1 points, about twice the standard deviation of a difference between two single runs, while
removing direction, count or distractors changes it by at most 0.3 points. Every variant retains between 78\% and 95\% of the gain on TVBench and TempCompass. Apart from the speed-change effect on velocity questions, the improvement is therefore spread across the
curriculum. The change in MotionBench repetition-count accuracy ranges from $-10.7$ to $-6.3$ points across all variants, including those without the counting or number-of-movers primitives, so the decline is not caused by the counting content alone.

\section{Hyper-parameters and replay}
\label{sec:hparam}

\cref{tab:hparam} varies one setting at a time. Self-distillation replay is the only component whose removal changes
the picture. Without replay the gain on TOMATO, TVBench and TempCompass is unchanged, but MotionBench drops by 8.8 points
instead of 1.9 and MVBench by 1.1 instead of rising by 2.1, so replay protects the model's behaviour on real videos
without limiting what it learns from the curriculum. Replay for half of the curriculum examples is as effective as replay for all of them. The other settings move the four-benchmark mean between +6.7 and +8.4 points against +7.9 for the paper
configuration, and every variant keeps a TOMATO gain of at least 17.9 points. A smaller learning rate mainly lowers the
TOMATO gain, a larger one costs MotionBench and MVBench accuracy, and the held-out curriculum accuracy grows with the
amount of training, from 73.0\% with 2{,}000 clips to 84.2\% with 8{,}000. The rank of the adapters matters little, as ranks 8 and 32 give
means of +7.7 and +7.3 points against +7.9 for rank 16.

\section{Subspace ablation}
\label{sec:causal}

A linearly decodable direction need not be used by the model for its
prediction~\cite{elazar2021amnesic}. We therefore remove decodable subspaces from the residual stream
at inference. Using one half of the source videos, we fit a subspace of $k \in \{1, 4, 16\}$ directions at each probed layer in the second half of the language model by iterative nullspace projection~\cite{ravfogel2020null}. We fit a logistic probe, project out its direction and repeat. On the other half, we project the
last-token state of every block from the middle of the network onward onto the orthogonal complement of the subspace
of the nearest probed layer. Random subspaces of the same size serve as controls. The two halves are then swapped, so every question is evaluated once.

As \cref{tab:ablate} shows, the forced-choice accuracy remains nearly unchanged under ablation on all three backbones. Removing a subspace of 1 to 16
directions changes the forced-choice accuracy by at most 1.2 points (two of 164 questions, Molmo2 with $k{=}1$) and the
16-dimensional subspace by at most 0.6 points (one question), while two random subspaces of each size change it by at most 0.6
points, of which \cref{tab:ablate} lists the 16-dimensional ones.
The decodable signal is therefore not the sole carrier of the decision. This is consistent with redundant or nonlinear encoding, and the two regimes describe where the information becomes linearly available.

\begin{table}[t]
  \centering
  \small
  \caption{\textbf{Subspace ablation} at the last token from the middle layer onward: forced-choice accuracy (\%) on
  the 164 TOMATO speed-change questions. Random: range over two random 16-dimensional subspaces.}
  \label{tab:ablate}
  \setlength{\tabcolsep}{4pt}
  \adjustbox{max width=\linewidth}{\begin{tabular}{@{}lccccc@{}}
    \toprule
    & None & $k{=}1$ & $k{=}4$ & $k{=}16$ & Random ($k{=}16$) \\
    \midrule
    Qwen3.5-9B & 90.2 & 90.2 & 90.9 & 90.9 & 90.2--90.2 \\
Qwen3-VL-8B & 84.1 & 83.5 & 84.1 & 84.1 & 84.1--84.8 \\
Molmo2-8B & 90.2 & 89.0 & 90.2 & 90.2 & 89.6--90.2 \\
 
    \bottomrule
  \end{tabular}}
\end{table}

\end{document}